\documentclass[journal]{IEEEtran}

\usepackage{xcolor, soul, framed}

\colorlet{shadecolor}{yellow}
\usepackage[pdftex]{graphicx}

\usepackage[cmex10]{amsmath}
\usepackage{array}
\usepackage{mdwmath}
\usepackage{mdwtab}
\usepackage{eqparbox}
\usepackage{url}
\usepackage{hyperref}
\usepackage[utf8]{inputenc}
\usepackage{amssymb}

\usepackage{algorithm}
\usepackage{algpseudocode}

\usepackage{comment}
\usepackage{todonotes}

\usepackage{placeins}

\begin{document}

\title{An approach to robust ICP initialization}

\author{Alexander Kolpakov and Michael Werman
\thanks{A. Kolpakov is at the
Universit\'e de Neuch\^atel. This work was funded in part by the Swiss National Science Foundation [grant no. PP00P2--202667]}
\thanks{M. Werman is at The Hebrew University of Jerusalem. This work was funded in part by the Israeli Science Foundation}
}

\maketitle

\begin{abstract}
    In this note, we propose an approach to initialize the Iterative Closest Point (ICP) algorithm  to match unlabelled point clouds   related by rigid transformations. 
    The method is based on matching the ellipsoids defined by the points' covariance matrices and then testing the various principal half--axes matchings that differ by elements of a finite reflection group.   
    We derive bounds on  the robustness of our approach to noise and numerical experiments  confirm our theoretical findings. 
\end{abstract}


\section{Introduction}

Point set registration is 
 aligning two point clouds using a rigid transformation. The purpose of finding such a transformation includes merging multiple data sets into a globally consistent
coordinate frame,  mapping a new measurement to a known data set to identify features, or finding the most similar object in a database, \cite{app,tut}.

Typically, 3D point cloud data are obtained from stereo, LiDAR, and RGB--D cameras, while 2D point sets are often extracted from features found in images.

Among the numerous registration methods proposed in the literature, the Iterative Closest Point (ICP) algorithm \cite{121791,132043,enwiki:1095051621}, introduced in the early 1990s, is the main algorithm for  registering  2D or 3D point sets using a rigid transformation. 
 
The Iterative Closest Point algorithm contrasts with the Kabsch \cite{Kabsch:a12999} algorithm and other solutions to the orthogonal Procrustes problem in that the Kabsch algorithm requires a correspondence between the point sets as input, whereas ICP treats correspondence as a variable to be estimated,
labeled vs. unlabeled.

The main ICP steps are:

\begin{algorithm}
\caption{ICP}\label{alg:icp}
\begin{algorithmic}

\While{not termination condition}
    \State Find a partial mapping between the sets
    \State Estimate the transformation 
based on  the previous step
    \State Transform the source points 
\EndWhile
\end{algorithmic}
\end{algorithm}

As this is a highly non--convex problem it  only finds a good solution when the point sets are initially  close to being aligned.

Yang et al.
\cite{yang2015go} present a branch--and--bound scheme for searching the entire 3D rigid transformation space, thus guaranteeing a global optimum. An overview of transformation estimation between  point clouds is presented in \cite{huang2021comprehensive,app}, including optimization--based and deep learning methods which can also be used to initialize the ICP in certain situations.
Several papers 
\cite{serafin2015nicp,SERAFIN201791,Makovetskii}, propose using local information and not just the points' coordinates to improve the matching.

This paper presents a simple and provably robust approach to preregister point sets before applying the ICP algorithm.
The method is based on matching the ellipsoids defined by the points' covariance matrices and testing the various principal half--axes matchings that differ by elements of a finite reflection group.   

\section{Main algorithm}

Given $n$ points in $\mathbb{R}^{d}$ with at least $d+1$  of them distinct we represent it as a $d \times n$ matrix.
The matrix is unique up to a permutation of the columns, or equivalently by right multiplication with a permutation matrix.

Let $\mathrm{O}(d)$ denote the orthogonal matrix group acting on $\mathbb{R}^d$, and $\mathrm{Sym}(n)$  the group of $n\times n$ permutation matrices. 

Given a point cloud $X$  with $n$ points, let $b(X)$ denote the barycenter of $X$, 
$$b(X) = \frac{1}{n} \sum^n_{i=1} x_i$$
and
$$\overline{X} = X-\frac{1}{n}X\mathbf{1}_{n \times n}=X - b(X)\,\mathbf{1}_{1 \times n}$$ 
and $\overline{X}$ the point set translated to be centered at the origin.

Given a point cloud $X \in \mathbb{R}^{d\times n}$, let $E(X) = X \, X^t \in \mathbb{R}^{d\times d}$, denote its inertia ellipsoid, also known as the covariance matrix. 
Note that $E(X)$ is  positive semidefinite  for any $X$. Generically  $\mathrm{rank}\, X = d$ and defines an ellipsoid with non--zero principal half--axes. 
Otherwise, we have a cylinder in $\mathbb{R}^d$ whose cross--section is a lower--dimensional ellipsoid. However, we shall only consider the generic case when $\mathrm{rank}\, X = d$, and thus the point cloud cannot be placed inside a lower--dimensional subspace of $\mathbb{R}^d$. In this case, $E = E(X) \in \mathbb{R}^{d\times d}$ is a positive definite matrix. 

Let $E \in \mathbb{R}^{d\times d}$ be a positive definite matrix with  distinct eigenvalues $\lambda_1 > \lambda_2 > \ldots > \lambda_d > 0$. Our next observation is that there are only finitely many ways to  orthogonally  diagonalize  $E$. 

Let $\Lambda = \mathrm{diag}(\lambda_1, \ldots, \lambda_d)$. Let $E = U_1 \Lambda U^t_1 = U_2 \Lambda U^t_2$, for two matrices $U_1,\, U_2 \in \mathrm{O}(d)$. Then 
$U^t_2 U_1 \in \mathrm{O}(d)$ is a self--isometry of the ``canonical'' ellipsoid $E_0$ defined by the equation $\lambda_1 x^2_1 + \ldots + \lambda_d x^2_d = 1$ having $\frac{1}{\lambda_1}, \ldots, \frac{1}{\lambda_d}$ as the lengths of its half--axes. Provided the above distinctness condition on $\lambda$'s, the only self--isometries of $E_0$ are reflections in the coordinate hyperplanes and their compositions. 

Let $\mathrm{Ref}(d)$ be the group generated by hyperplane reflections of the form $x_i \mapsto - x_i$ for  $i \in \{1, 2, \ldots, d\}$, $x_j \mapsto x_j$, for $j \neq i$, for all $i=1, 2, \ldots, d$. Note that $\mathrm{Ref}(d)$ is a finite group with $2^d$ elements. It consists of diagonal matrices with $+1$'s and $-1$'s on the diagonal.

Thus, if $E$ can be brought to the same diagonal form by using two orthogonal matrices $U_1, U_2 \in \mathrm{O}(d)$, then $U^t_1 U_2 \in \mathrm{Ref}(d)$. This means that there are no more than $2^d$ ways to diagonalize $E$  using orthogonal matrices. 

Assume that two point clouds $P, Q \in \mathbb{R}^{d\times n}$ are related by a distance--preserving transformation, that is $Q$ is obtained from $P$ by applying an orthogonal transformation followed by a translation. 

Let $O \in \mathrm{O}(d)$ be an orthogonal transformation, and $S \in \mathrm{Sym}(n)$ a permutation matrix. For any point cloud $X \in \mathbb{R}^{d\times n}$ we have  $b(O \, X) = O\, b(X)$ and $b(X\, S) = b(X)$. 

Given a point cloud $X\in \mathbb{R}^{d\times n}$ we  assume that $\mathrm{rank}\,X = d$ and $E(X)$ has simple spectrum (all its eigenvalues are distinct). This assumption is generically true (see \cite{arnold72} for more details on spectra of symmetric matrices), and can always be achieved by a small perturbation of the points in $X$. 

As noted by many, when minimizing the least square distance between point sets their barycenters should be aligned, following from 
$$
\frac{\partial \sum^{n}_{i=1} ||Ox_i + b -y_i||^2}{\partial b} = 0 
$$
implies
$$
b = \frac{1}{n}\sum^{n}_{i=1} y_i-\frac{1}{n}\sum^{n}_{i=1} Ox_i. 
$$
Thus, we may assume that $P$ and $Q$ are replaced with $\overline{P}$ and $\overline{Q}$, centered at the origin. Then  $Q = O\, P \, S$, for an orthogonal transformation $O \in \mathrm{O}(d)$ and a permutation matrix $S\in \mathrm{Sym}(n)$. 

Moreover, let $E_X=E(X)$, then  $E_Q = O\, E_P \, O^t$ independent of the permutation $S$, ($S^{-1}=S^t$). This also means that $E_P$ and $E_Q$ have the same set of eigenvalues, which are  distinct by the previous assumption. 

We  use the following algorithm, {\it E--Init} for ICP initialization. 

\begin{algorithm}
\caption{Ellipsoid Initialization ({\it E--Init})}\label{alg:init}
\begin{algorithmic}
    \State $P \leftarrow \overline{P}$, $Q \leftarrow \overline{Q}$ 
    \Comment{centering}
    \State $E_P \leftarrow  P \, P^t$, $E_Q \leftarrow Q \, Q^t$.
    \State  $E_P \leftarrow U_P \, \Lambda \, U^t_P$, $E_Q \leftarrow U_Q\, \Lambda \, U^t_Q$ \Comment{diagonalization}
    \State $U_0 \leftarrow U_Q \, U^t_P$
    \State  $D^* \leftarrow 
    \underset{D \in \mathrm{Ref}(d)}
    {\mathrm{argmin}}\text{Match}(P, (U_0 \, U_P \, D \, U^t_P)^t\, Q)$
    \State
    $U \leftarrow U_0 \, U_P \, D^* \, U^t_P$ \Comment{initial rigid motion}
    \State
    $Q\leftarrow U^t\,Q$
\end{algorithmic}
\end{algorithm}

$\text{Match}(X, Y)$ provides the nearest neighbor matching distance of two sets $X, Y \in \mathbb{R}^{d \times n}$ (e.g. based on $k$-$d$--trees). After {\it E--Init}, run ICP on $P$ and $Q$.

\section{Correctness of the algorithm}

Even under  ideal conditions, there is an obstacle to recovering the original $O$ and $S$. Namely, the cloud $P$ (and thus $Q$) may have symmetries: an orthogonal transformation $O' \in \mathrm{O}(d)$ is a symmetry of a point cloud $X \in \mathbb{R}^{d\times n}$ if there exists a permutation $S' \in \mathrm{Sym}(n)$ such that $O' \, X = X \, S'$. Then $U = O O'$, $M = (S')^t S$ will also be a perfect matching of the points of $P$ to the points of $Q$. 

In what follows we assume that $P$ and $Q$ have no symmetries, which can be achieved generically by a slight perturbation of the respective point clouds. This assumption is not restrictive: in the presence of symmetries, we will have multiple solutions, each of which is as good as any other. 

The previous discussion was for {\it exact} alignment 
of {\it equal} sized point sets, the following sections show that both these restrictions can be relaxed.

\section{Robustness to noise}\label{noise}

\subsection{Multiplicative noise}\label{noise-mult}

We first  consider  multiplicative noise. Noise of this type changes its magnitude depending on the size of point clouds. One natural example is \textit{relative} measurement errors.  

Let $N \in \mathbb{R}^{d\times n}$ be a matrix with entries $N_{ij} \sim \mathcal{N}(1, \sigma^2)$, $i = 1, \ldots, d$, $j = 1, \ldots, n$, independent Gaussian random variables and $\odot$ denote the Hadamard (elementwise) product of matrices.   $\mathbb{E}$ and $\mathbb{P}$  denote the expectation (componentwise for matrices) and probability. 

We  model a noisy point cloud $Q = O\, P\, S$ by masking $Q$ with $N$, i.e. replacing $Q$ with $Q' = Q \odot N$. In this case, we have  $\mathbb{E}[Q'] = Q$ and, moreover, if $b(Q) = 0$ then $\mathbb{E}[b(Q')] = 0$, too.

Let $ \mathbb{E}_{Q'}=\mathbb{E}[E(Q')] = \mathbb{E}[E(Q\odot N)]$, 
$E_{Q'} = E(Q') = E(Q \odot N)$ and $E_Q = E(Q)$ be the corresponding ellipsoids: the first being the ``average ellipsoid'' (averaged over the noise,  $N$), the second being the ``noisy'' one, and the third being the ``ideal'' one. 

First we compare $ \mathbb{E}_{Q'} $ to $E_Q$. By a straightforward computation, we obtain that 
\begin{multline}
    \mathbb{E}_{Q'}  = \mathbb{E}[(Q\odot N)(Q\odot N)^t] = (Q\, Q^t) \odot (\mathbf{1}_{d}\,\mathbf{1}^t_{d} + \Sigma) = \\ = E_Q + \sigma^2\, \mathrm{diag}\,E_Q,
\end{multline}
where $\Sigma = \mathrm{diag}(\sigma^2, \ldots, \sigma^2)$.

Thus,
\begin{equation}\label{diff1}
    \|\mathbb{E}_{Q'}  - E_Q \|_2 = \sigma^2 \max_{i=1..d} E_{ii} \leq \sigma^2\, \|E_Q\|_2. 
\end{equation}

In particular, this means that the ``average'' ellipsoid $\mathbb{E}_{Q'}$ does not deviate from $E_Q$ too much  if the  relative noise is small enough.  

The above applies to any noise matrix $N = (N_{ij}) \in \mathbb{R}^{d\times n}$ where the entries $N_{ij}$ are i.i.d. random variables with $\mathbb{E}[N_{ij}] = 1$ and $\mathrm{Var}[N_{ij}] = \sigma^2$. 

Using the analog of Chebyshev's inequality for symmetric positive definite matrices (see \cite[Theorem 14]{Ahlswede:2002})] 
\begin{equation}\label{prob1}
    \mathbb{P}\left( \| E_{Q'} - \mathbb{E}_{Q'} \|_2 > \delta \right) \leq \frac{\mathrm{tr}\, S^2}{\delta^2}
\end{equation}
where $S^2 = \mathbb{E}[(E_{Q'})^2] - (\mathbb{E}[E_{Q'}])^2$ is the variance matrix (computed elementwise) and $\delta > 0$ is a constant.

We have that
\begin{equation}\label{tr1}
    \mathrm{tr}\, \mathbb{E}[(E_{Q'})^2] = \mathbb{E}[\mathrm{tr}\,(E_{Q'})^2] \leq \mu'_4\, \mathrm{tr}\,E^2_Q
\end{equation}
where the inequality is obtained by  assuming that $N = (N_{ij}) \in \mathbb{R}^{d\times n}$ has independent entries and $\mu'_4 \geq \mu'_3 \mu'_1$, $\mu'_4 \geq (\mu'_2)^2$, where $\mu'_l$ is the $l$--th (non--central) moment of $\mathcal{N}(1, \sigma^2)$, i.e. $\mu'_1 = 1$, $\mu'_2 = 1 + \sigma^2$, $\mu'_3 = 1 + 3 \sigma^2$, $\mu'_4 = 1 + 6 \sigma^2 + 3 \sigma^4$. 

Also, we have 
\begin{multline}
    \label{tr2}
    \mathrm{tr}\,(\mathbb{E}[E_{Q'}])^2 = \mathrm{tr}\,(E_Q + \sigma^2\,\mathrm{diag}\,E_Q)^2 = \\ = \mathrm{tr}\,(E^2_Q) + 2\sigma^2 \Delta + \sigma^4 \Delta,
\end{multline}
where $\Delta = \sum^d_{i=1}E_Q[i,i]^2$.

We  estimate $\mathrm{tr}\,S^2$ by combining Eq. \ref{tr1} and Eq. \ref{tr2} as follows:
\begin{multline}
    \label{tr3}
    \mathrm{tr}\,S^2 = \mathrm{tr}\,\mathbb{E}[(E_{Q'})^2] - \mathrm{tr}\,(\mathbb{E}[E_{Q'}])^2 \\ \leq \sigma^2\, (2 + \sigma^2)\, (3\, \mathrm{tr}\,E^2_Q - \Delta) \\ \leq 9\, d\, \sigma^2\, \|E_Q\|^2_2,
\end{multline}
as we assume $\sigma \leq 1$. 

Finally, by combining Eq. \ref{prob1} with Eq. \ref{tr3}, we obtain
\begin{equation}\label{prob2}
    \mathbb{P}\left( \| E_{Q'} - \mathbb{E}_{Q'} \|_2 > \delta \right) \leq \frac{9\, d\, \sigma^2\, \| E_Q \|^2_2}{\delta^2}.
\end{equation}

Setting $\delta = \varepsilon\, \|E_Q\|_2$ in Eq. \ref{prob2},
\begin{equation}\label{prob3}
    \mathbb{P}\left( \| E_{Q'} - \mathbb{E}_{Q'} \|_2 > \varepsilon\, \| E_Q \|_2 \right) \leq d\, \left( \frac{3\sigma}{\varepsilon} \right)^2,
\end{equation}
for any fixed $\varepsilon > 0$. In this instance we can fulfill the usual ``three--sigma'' rule by choosing $\varepsilon = \sqrt{3 d \sigma}$. 

Thus, Eq. \ref{diff1} and Eq. \ref{prob3} imply that with probability $1 - \Omega(\sigma)$  we have
\begin{multline}\label{diff2}
    \| E_{Q'} - E_Q \|_2 \leq \| E_{Q'} - \widetilde{E}_Q \|_2 + \| \widetilde{E}_Q - E_Q \|_2 \\ \leq (\sqrt{3\,d\,\sigma} + \sigma^2) \, \| E_Q \|_2 = (\sqrt{3\,d\,\sigma} + o(\sigma)) \, \| E_Q \|_2.
\end{multline}

Note that above we measure the magnitude of the scale--invariant quantity $\frac{\| E_{Q'} - E_Q \|_2}{\| E_Q \|_2}$ that corresponds to the relative difference of the ellipsoids. Indeed, simultaneous scaling of the point clouds by a non--zero factor   induces the same scaling of the ellipsoids. However, this should  not affect  the orthogonal transformations or the point matching between the clouds. 

Let $U$ be obtained by running {\it E--Init} on $P$ and $Q'$ as input: we have $U = \mathrm{argmin}_{U \in \mathrm{O}(d)}\, \| U\, E_P\, U - E_{Q'} \|_2$. Thus, 
\begin{multline}\label{norm-estimate-mult}
        \| E_P - U^t\,O\,E_P\,O^t\,U\|_2 = \| U\,E_P\, U^t - O\, E_P O^t \|_2 \\ 
        \leq \| U\, E_P\, U^t - E_{Q'}\|_2 + \| E_{Q'} - E_Q \|_2 \\ 
        \leq \| O\, E_P\, O^t - E_{Q'}\|_2 + \| E_{Q'} - E_Q \|_2 = \\
        = 2\, \| E_{Q'} - E_Q \|_2 \leq (\sqrt{12\,d\,\sigma} + o(\sigma))\, \| E_Q \|_2.
\end{multline}

Since $\| E_Q \|_2 = \| E_P \|_2$, we can rewrite the above as 
\begin{equation}\label{diff3}
    \| \widehat{E}_P - U^t\,O\,\widehat{E}_P\,O^t\,U\|_2  \leq \sqrt{12\,d\,\sigma} + o(\sigma),
\end{equation}
where $\widehat{E}_P = \frac{E_P}{\| E_P \|_2}$ is the ``normalized'' ellipsoid. As remarked above, such normalization does not affect our analysis of the proximity of $O$ and $U$. 

Once $\| \widehat{E}_P - W\, \widehat{E}_P\, W^t \|_2 = 0$ for a $W \in \mathrm{O}(d)$ we have  $W\, \widehat{E}_P = \widehat{E}_P\, W$. Up to a change of basis we may assume that $E_P = \Lambda = \mathrm{diag}(\lambda_1, \ldots, \lambda_d)$ with $\lambda_1 > \lambda_2 > \ldots > \lambda_d > 0$ all distinct, as before. Then $\widehat{E}_P = \mathrm{diag}(1, \lambda^{-1}_1 \lambda_2, \ldots, \lambda^{-1}_1 \lambda_n)$. This implies immediately that $W$ is diagonal, and thus $W \in \mathrm{Ref}(d)$. Since by assumption $P$ does not have symmetries, we can only have $W$ being the identity transformation $I_d$.

Thus, if $\varepsilon$ and $\sigma$ in Eq. \ref{diff3} are small enough, we have  $U^t\, O$ is close to $I_d$, and thus $U$ that matches the specimen cloud $P$ to the noisy cloud $Q'$ will approximate well the original transformation $O$ that matches $P$ exactly to $Q$.

\subsection{Additive noise}

Additive noise is also present under usual circumstances. One example is the background noise in the communication channel. 

In this case, let $N \in \mathbb{R}^{d\times n}$ be a matrix with entries $N_{ij} \sim \mathcal{N}(0, \sigma^2)$, $i = 1, \ldots, d$, $j = 1, \ldots, n$, independent Gaussian random variables. To model noise, we replace $Q$ with $Q' = Q + N$. As before, $\mathbb{E}[Q'] = Q$. Moreover, once $b(Q) = 0$, then $\mathbb{E}[b(Q')] = 0$, too. 

Let $\xi = \| E_{Q'} - E_Q  \|_F$ be the non--negative random variable measuring the difference between the ``noisy'' ellipsoid $E_{Q'}$ and the ``ideal'' ellipsoid $E_Q$. Here we prefer to use the Frobenius norm instead of the spectral radius, although these norms are equivalent and our choice is only that of convenience. 

First, we compute
\begin{equation}
    E_{Q'} = (Q + N)\,(Q + N)^t = E_Q + Q\, N^t + N\, Q^t + N\,N^t,
\end{equation}
and thus
\begin{equation}
    \xi = \| E_{Q'} - E_Q  \|_F \leq 2 \, \|Q\|_F \, \|N\|_F + \|N\|^2_F,
\end{equation}
by subadditivity and submultiplicativity of the Frobenius norm. 

As $\mathbb{E}[\sqrt{\eta}] \leq \sqrt{\mathbb{E}[\eta]}$ for any non--negative random variable, we obtain
\begin{multline}
    \mathbb{E}[\xi] \leq 2 \, \|Q\|_F \sqrt{\mathbb{E}[\| N \|^2_F]} + \mathbb{E}[\| N \|^2_F] = \\
    = 2 \, \|Q\|_F \, \sqrt{nd}\, \sigma + n\,d\,\sigma^2. 
\end{multline}

As $\| X + Y \|^2_F \leq 2\, \| X \|^2_F + 2\, \| Y \|^2_F$ for any $X, Y \in \mathbb{R}^{d \times d}$, we get
\begin{multline}
    \xi^2 = \| Q\, N^t + N\, Q^t + N\, N^t \|^2_F \\ \leq 8\,\| Q \|^2_F \, \| N \|^2_F + 2\, \| N \|^4_F.
\end{multline}
Here we also use submultiplicativity of the Frobenius norm. 

As noted before $\mathbb{E}[\| N \|^2_F = n\, d\, \sigma^2$. Also, we have $\mathbb{E}[\| N \|^4_F \leq n^2 d^2 \mu_4$, where $\mu_i$ is the $i$-th central moment of $\mathcal{N}(0, \sigma^2)$. The latter inequality follows from $\mu_4 = 3 \sigma^4 > \sigma^4 = \mu^2_2$. Thus
\begin{equation}\label{xi2-estim}
    \mathbb{E}[\xi^2] \leq 8\, \| Q \|^2_F\, n\, d\, \sigma^2 + 6\, n^2\, d^2\, \sigma^4.
\end{equation}

Let $\delta > 0$ be a positive real number. Then the classical Markov inequality implies
\begin{equation}\label{prob-estim1}
    \mathbb{P}\left( \|E_{Q'} - E_Q\|_F > \delta \right) \leq \frac{\mathbb{E}[\xi^2]}{\delta^2}.
\end{equation}

By setting $\delta = \sqrt{3 \sigma} \| Q \|_F$ in Eq. (\ref{prob-estim1}) and using Eq. (\ref{xi2-estim}) we compute
\begin{multline}\label{xi2-prob}
    \mathbb{P}\left( \|E_{Q'} - E_Q\|_F > \sqrt{3 \sigma}\, \| Q \|_F \right) \\ \leq \frac{8\, n\, d\, \sigma^2\, \|Q\|^2_F}{3\, \sigma\, \|Q\|^2_F} + \frac{6\, n^2\, d^2\, \sigma^4}{3\, \sigma\, \|Q\|^2_F} \leq 3\, n\, d\, \sigma + o(\sigma).
\end{multline}

Let $\widetilde{E}_P = \frac{E_P}{\| P \|_F}$ be the ``normalized'' ellipsoid. From Eq. (\ref{xi2-prob}), and by applying an estimate analogous to Eq.~(\ref{norm-estimate-mult})  we obtain that
\begin{equation}\label{diff-estim}
    \| \widehat{E}_P - U^t\, O\, \widehat{E}_P\, O^t\, U \|_F \leq \sqrt{12\, \sigma}
\end{equation}
holds with probability $1 - \Omega(\sigma)$.

Using Eq.~\ref{diff-estim} and applying an analogous argument to that in  Subsection~\ref{noise-mult}, we obtain that for sufficiently small $\sigma$ the transformation $U^t\,O$ is close to the identity transformation $I_d$. Thus, the recovered transformation $U$ that matches the specimen cloud $P$ to the noisy cloud $Q'$ will approximate $O$ well enough. 

\section{Number of points discrepancy}\label{occlusion}

Now let $Q = O\, P\, S$, for a point cloud $P \in \mathbb{R}^{d\times k}$, an orthogonal transformation $O \in \mathrm{O}(d)$ and a permutation matrix $S\in \mathrm{Sym}(k)$ but in this case, we are given two point clouds 
$P'$ and $Q'$ 
such that $P' \supset P$ and  $Q' \supset Q$.

We assume that the two point clouds $P'$ and $Q'$ can be partially matched  using ICP despite  having different cardinalities. We need to understand how the norms of $\Delta P = P'\setminus P$ and $\Delta Q = Q' \setminus Q$  affect our approach. 

Let $E_{P'} = E(P')$, $E_{Q'} = E(Q')$, $ E_{\Delta P} = E(\Delta P)$, and $E_{\Delta Q} = E(\Delta Q)$. A straightforward computation shows that $E_{P'} = E_P + E_{\Delta P} $ and $E_{Q'} = E_Q +  E_{\Delta Q}$. Note that  $E_Q = O\, E_P\, O^t$. 

Let $U \in \mathrm{O}(d)$ be obtained by running {\it E--Init} with point clouds $P'$ and $Q'$ as input. Then we obtain that $U = \mathrm{argmin}_{U \in \mathrm{O}(d)}\,\| U\, E_{P'}\, U^t - E_{Q'} \|_2$. 

Thus, we get
\begin{multline}
    \| E_P - U^t\, O\, E_P\, O^t\, U  \|_2 = \| E_P - U^t\, E_Q\, U \|_2 \leq \\ 
    \leq \| E_P - E_{P'} \|_2 + \| E_{P'} - U^t\, E_{Q'}\, U \|_2 + \\ + \| U^t\, E_{Q'}\, U - U^t\, E_Q\, U \|_2
    = \| E_{\Delta P} \|_2 + \| E_{\Delta Q} \|_2 + \\
    + \| E_{P'} - U^t\, E_{Q'}\, U \|_2 \leq \| E_{\Delta P} \|_2 + \| E_{\Delta Q} \|_2 +\\
    + \| E_{P'} - O^t\, E_{Q'}\, O \|_2 \leq \| E_{\Delta P} \|_2 + \| E_{\Delta Q} \|_2 +\\
    + \| E_{\Delta P} - O^t\,E_{\Delta Q}\,O \|_2 \leq 2 \left( \| E_{\Delta P} \|_2 + \| E_{\Delta Q} \|_2 \right).
\end{multline}

Assume that $\|E_{\Delta P}\|_F \leq \frac{\varepsilon}{4\sqrt{d}}\,\| E_P \|_F$ and $\|E_{\Delta Q}\|_F \leq \frac{\varepsilon}{4\sqrt{d}}\,\| E_Q \|_F$. This will be the case, e.g. when the norm of vectors in $P$ and $\Delta P$ are relatively comparable while $P$ contains considerably more points than $\Delta P$ and the same holds for $Q$ and $\Delta Q$. 

For any non--degenerate matrix $X \in \mathbb{R}^{d\times d}$ we have   $\|X\|_2 \leq \|X\|_F \leq \sqrt{d}\, \|X\|_2$ between its Frobenius norm $\|X\|_F$ and its spectral radius $\|X\|_2$. Thus we get
\begin{multline}
    \| E_P - U^t\,O\, E_P\, O^t\,U \|_2 \\ \leq 2\,\frac{\varepsilon}{4}\, ( \| E_P \|_2 + \| E_Q \|_2 ) \leq \varepsilon\,\| E_P \|_2. 
\end{multline}

As before, for the normalized ellipsoid $\widehat{E}_P$ we obtain 
\begin{equation}
    \| \widehat{E}_P - U^t\,O\, \widehat{E}_P\, O^t\,U \|_2 \leq \varepsilon. 
\end{equation}

Once $\varepsilon$ is small enough, the same logic as in Section~\ref{noise} leads to the conclusion that, up to a change of basis, we have $U = O\, D$ for an element $D \in \mathrm{Ref}(d)$. Since in Step 4 of the  algorithm each element of $\mathrm{Ref}(d)$ is tested, we still shall obtain the best match starting with $U$. The rest of the algorithm will then work out as usual. 

Thus, the transformation $O\in \mathrm{O}(d)$ and the point matching $S\in \mathrm{Sym}(k)$ will still 
be recovered once we initialize ICP  to match two  large parts of $P \in \mathbb{R}^{d\times k}$ of a point cloud $P' \in \mathbb{R}^{d\times m}$ and $Q = O\,P\,S \in \mathbb{R}^{d\times k}$ of a point cloud $Q' \in \mathbb{R}^{d\times n}$. 

\section{Superposition of errors}

If several of the above issues (multiplicative or additive noise, or   point cardinality  difference) are present, then the errors  add  according to the triangle inequality for matrix norms. If all three kinds of errors are relatively small, our algorithm is still robust. We provide experimental evidence below, Subsection~\ref{super}. 
%The SageMath worksheet used to generate and perform the respective numerical experiemnts are available on GitHub \cite{github:2022}.   

\section{Possible modifications}

The algorithm is based on aligning the eigenvectors (principal half--axes) of two ellipsoids associated with the respective points clouds. In the above, eigenvectors are  matched  based on the order of eigenvalues.
However, in the case of excessive noise,  eigenvalues may  switch. To tackle this issue, instead of treating the uncertainty in ellipsoid matching as an element of $\mathrm{Ref}(d)$, we assume that we can be wrong up to a bigger finite group: the $2^d \, d!$ element $B_d$ Coxeter group that also permutes the coordinate axes.

\section{Numerical experiments}
\subsection{Statistics definitions}
The following statistics were measured in the experiments to quantify and compare results.
\begin{itemize}
    \item the added noise (normalized): $\nu = \| Q' - Q \|_2 / \|P\|_2$;
    \item the normalized distance to the noisy image $Q'$: \\$\delta = \| Q' - Q_{icp} \|_2 / \|P\|_2$;
    \item the normalized distance to the actual (without noise) specimen image: $\delta_{spec} = \| Q - U_{icp} \, P\, S \|_2 / \|P\|_2$;
    \item the success rate of the test batch: $100$ tests with success criterion $\delta_{spec} \leq 0.05$;
    \item the distance to the original orthogonal transformation: \\$\delta_o = \| U_{icp} - O \|_2$;
    \item the normalized Hamming distance to the original permutation: $\delta_H = \| S_{icp} - S \|^2_F/(2 \cdot n)$;
    \item the ICP change in the (normalized) distance: \\$\delta^{icp} = \left( \| Q' - Q_{init} \|_2 - \| Q' - Q_{icp} \|_2 \right) / \|  P \|_2$;
    \item the ICP change to the orthogonal transformation: \\$\delta^{icp}_o = \| U_{init} - U_{icp} \|_2$.
\end{itemize}
\subsection{{\it E--Init}: random point clouds}

$100$ tests were performed. Each time a random point cloud $P \subset \mathbb{R}^3$ with $100$ points is generated. The points in $P$ are distributed uniformly in the cube $[-20, 20]^3$ (each coordinate being uniformly distributed). Then a random orthogonal matrix $O \in \mathrm{O}(3)$ and a random permutation $S \in \mathrm{Sym}(100)$ are generated, and the cloud $Q = O\, P\, S$ is produced. In all tests, ICP initialized with our algorithm   recovered $O$ and $S$ (up to machine precision). In fact, we only need to run ICP to recover the nearest neighbors, and thus the permutation $S$. The orthogonal transformation $U = U_0 U_P D^* U^t_P$ from  {\it E--Init}  is already equal to the original $O$ under ideal conditions. 

\subsection{{\it E--Init}: Caerbannog point clouds}

$100$ tests as described above were performed on each of the three unoccluded Caerbannog clouds \cite{caerbannog:2020}: the teapot, the bunny\footnote{Also known as the Killer Rabbit of Caerbannog.}, and the cow. All tests  successfully recovered both $O$ and $S$. 

\subsection{{\it E--Init}: comparison to ICP without intialization}

\begin{figure}[h]
\centering
\includegraphics[width=0.45\textwidth]{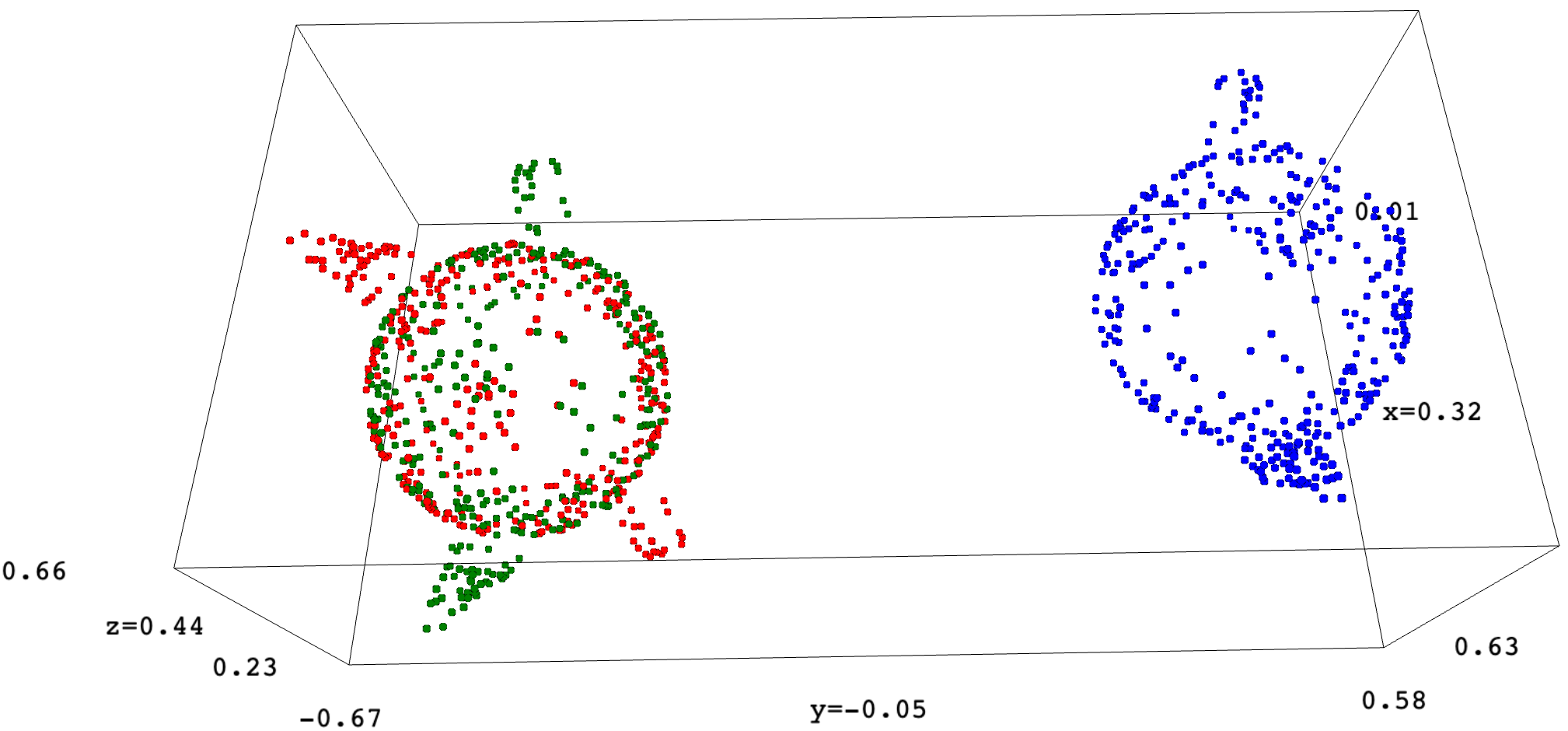}
\caption{A typical result of ICP applied to the ``teapot'' cloud without initialization: the specimen cloud $P$ (blue), the image $Q$ (red) of $P$ under an orthogonal transformation, and the image of $P$ recovered by ICP (green).}\label{teapot-no-init} 
\end{figure}

\begin{figure}[h]
\centering
\includegraphics[width=0.45\textwidth]{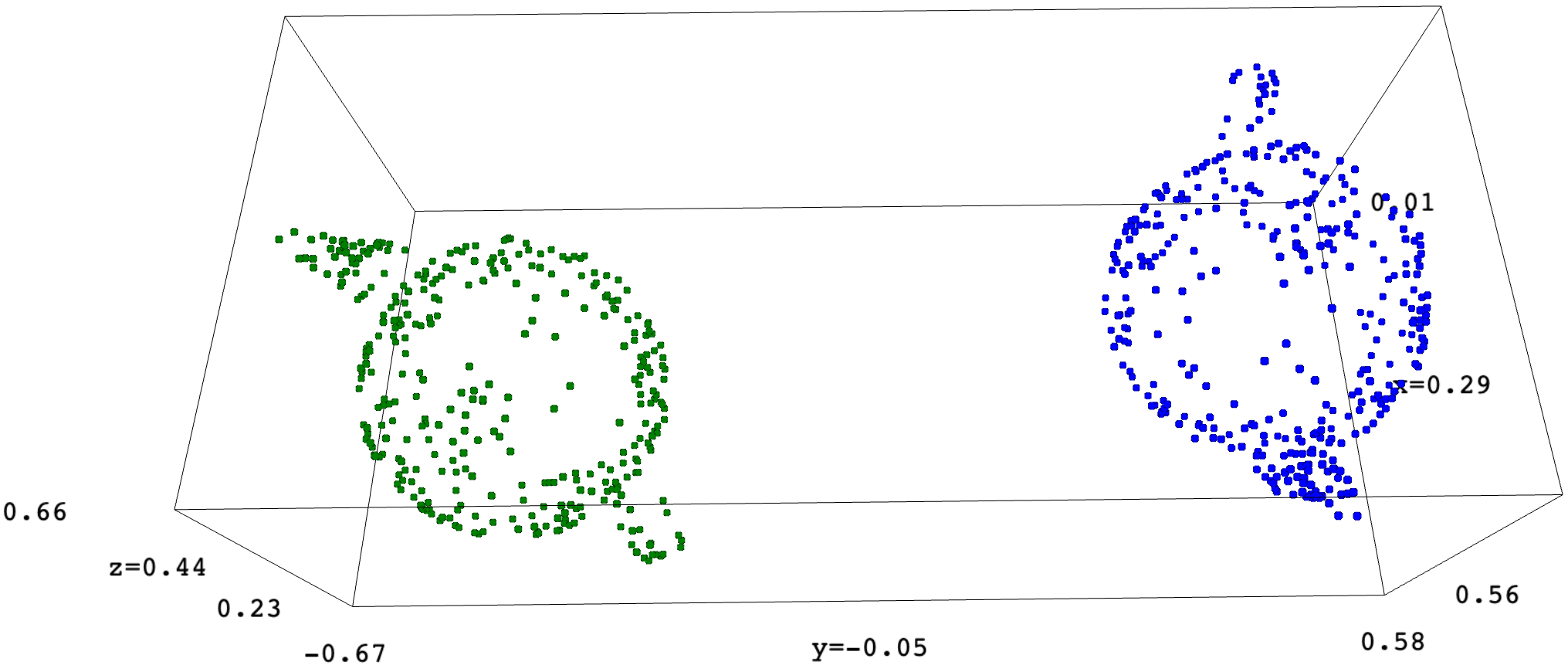}
\caption{A typical result ICP applied to the ``teapot'' cloud with {\it E--Init}. The specimen cloud $P$ is  in blue, while the image $Q$ of $P$ under an orthogonal transformation is red, and the image of $P$ recovered by ICP is green. The green and red images completely overlap. }\label{teapot-init} 
\end{figure}

We compare ICP initialized with {\it E--Init}
 to ICP without {\it E--Init} in the case of the Caerbannog clouds described above. In a batch of $100$ tests, we obtained (under ideal conditions) no higher than $2 \%$ success rate (i.e. $\geq 98 \%$ failures) without initialization and a $100\%$ success rate for ICP  with  {\it E--Init}. A sample of point cloud registration with and without is shown in Figures \ref{teapot-no-init} -- \ref{teapot-init}.

\subsection{Multiplicative noise}

$100$ tests were performed on each of the three unoccluded Caerbannog clouds \cite{caerbannog:2020}: the teapot, the bunny, and the cow. All tests were successful in recovering $O$ with relatively minor errors given that the noise is also relatively minor. Recovering the permutation $S$ fails in most cases as even relatively minor noise interferes seriously with the nearest neighbors matching. 

In each test, we obtain an orthogonal transformation $U_{init}$ from our initialization algorithm and then  an improved transformation $U_{icp}$ after running ICP. We also obtain the image of $P$ from ICP, i.e. $Q_{icp} = U_{icp} \, P\, S_{icp}$ and  compare it to $Q'$ after matching the nearest neighbors of $Q_{icp}$ and $Q'$  using a matching matrix $S_{icp}$. We  also compare $Q_{icp}$ to the actual specimen image $Q = O \, P\, S$ of $P$. 

We  find it instructive to compute the improvement that ICP makes to the initializing orthogonal transformation $U$, as well as the distance between $Q_{init} = U\,P\,S_{icp}$ and $Q_{icp}$. 

In the tests, we used Gaussian (multiplicative) noise $\mathcal{N}(1, \sigma^2)$ with $\sigma \in \{0.1, 0.2, \cdots, 0.6\}$. 
\begin{comment}
The following values were measured for each value of $\sigma$ in a batch of $100$ tests:

\begin{itemize}
    \item the added noise (normalized): $\nu = \| Q' - Q \|_2 / \|P\|_2$;
    \item the normalized distance to the noisy image $Q'$: \\$\delta = \| Q' - Q_{icp} \|_2 / \|P\|_2$;
    \item the normalized distance to the actual (without noise) specimen image: $\delta_{spec} = \| Q - U_{icp} \, P\, S \|_2 / \|P\|_2$;
    \item the success rate of the test batch: $100$ tests with success criterion $\delta_{spec} \leq 0.05$;
    \item the distance to the original orthogonal transformation: \\$\delta_o = \| U_{icp} - O \|_2$;
    \item the normalized Hamming distance to the original permutation: $\delta_H = \| S_{icp} - S \|^2_F/(2 \cdot n)$;
    \item the ICP change in the (normalized) distance: \\$\delta^{icp} = \left( \| Q' - Q_{init} \|_2 - \| Q' - Q_{icp} \|_2 \right) / \|  P \|_2$;
    \item the ICP change to the orthogonal transformation: \\$\delta^{icp}_o = \| U_{init} - U_{icp} \|_2$.
\end{itemize}
\end{comment}

\begin{figure}[t]
\centering
\includegraphics[width=0.33\textwidth]{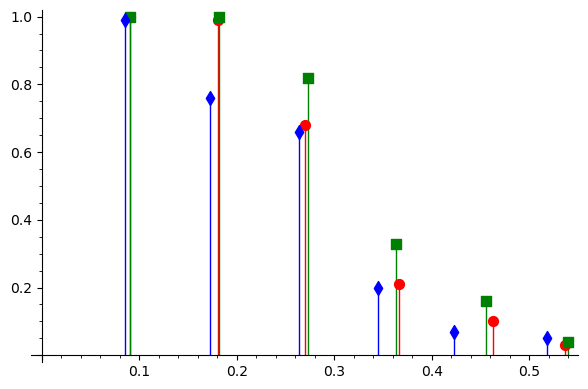}
\caption{Success rate (vertical axis-$\tau$) depending on the  multiplicative noise (horizontal axis-$\nu$)  over $100$ tests for  the ``teapot'' (red circle), ``bunny'' (green square), and ``cow'' (blue lozenge) point clouds. The experiments were carried out  for Gaussian multiplicative noise $\mathcal{N}(1, \sigma^2)$,  $\sigma \in (0.1, 0.2, \ldots, 0.6)$.}\label{success-rate-mul-noise}

\vspace{0.1in}

\centering
\includegraphics[width=0.33\textwidth]{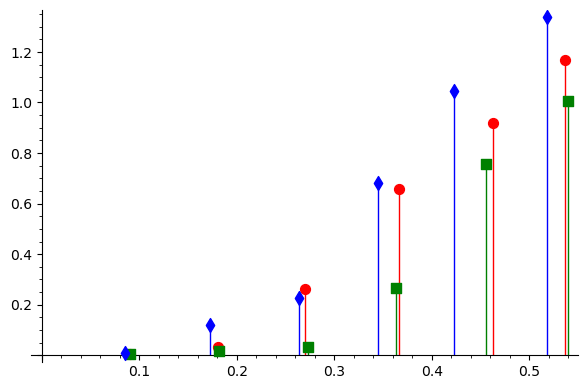}
\caption{Normalized distance to the specimen   depending on the  multiplicative noise  measured over $100$ tests for  the ``teapot'' (red circle), ``bunny'' (green square), and ``cow'' (blue lozenge) point clouds.  $\nu$ and  $\delta_{spec}$ are computed for Gaussian multiplicative noise $\mathcal{N}(1, \sigma^2)$,  $\sigma \in (0.1, 0.2, \ldots, 0.6)$.}\label{delta-spec-mul-noise}

\vspace{0.1in}

\centering
\includegraphics[width=0.33\textwidth]{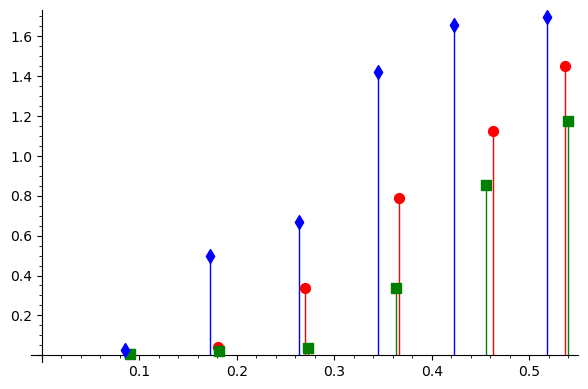}
\caption{Distance to the original orthogonal transformation   depending on the  multiplicative noise  measured over $100$ tests for each of the ``teapot'' (red circle), ``bunny'' (green square) and ``cow'' (blue lozenge) point clouds.  The experiments were carried out  for Gaussian  multiplicative noise $\mathcal{N}(1, \sigma^2),\sigma \in (0.1, 0.2, \ldots, 0.6$).}\label{delta-ortho-mul-noise}
\end{figure}

In Fig.~\ref{success-rate-mul-noise} we plot the success rate depending on the added noise $\nu$.  

Judging from Fig.~\ref{success-rate-mul-noise}, most tests fail for $\nu \leq 0.35$ (which approximately corresponds to $\sigma \geq 0.4$).

\begin{figure}[h]
\centering
\includegraphics[width=0.33\textwidth]{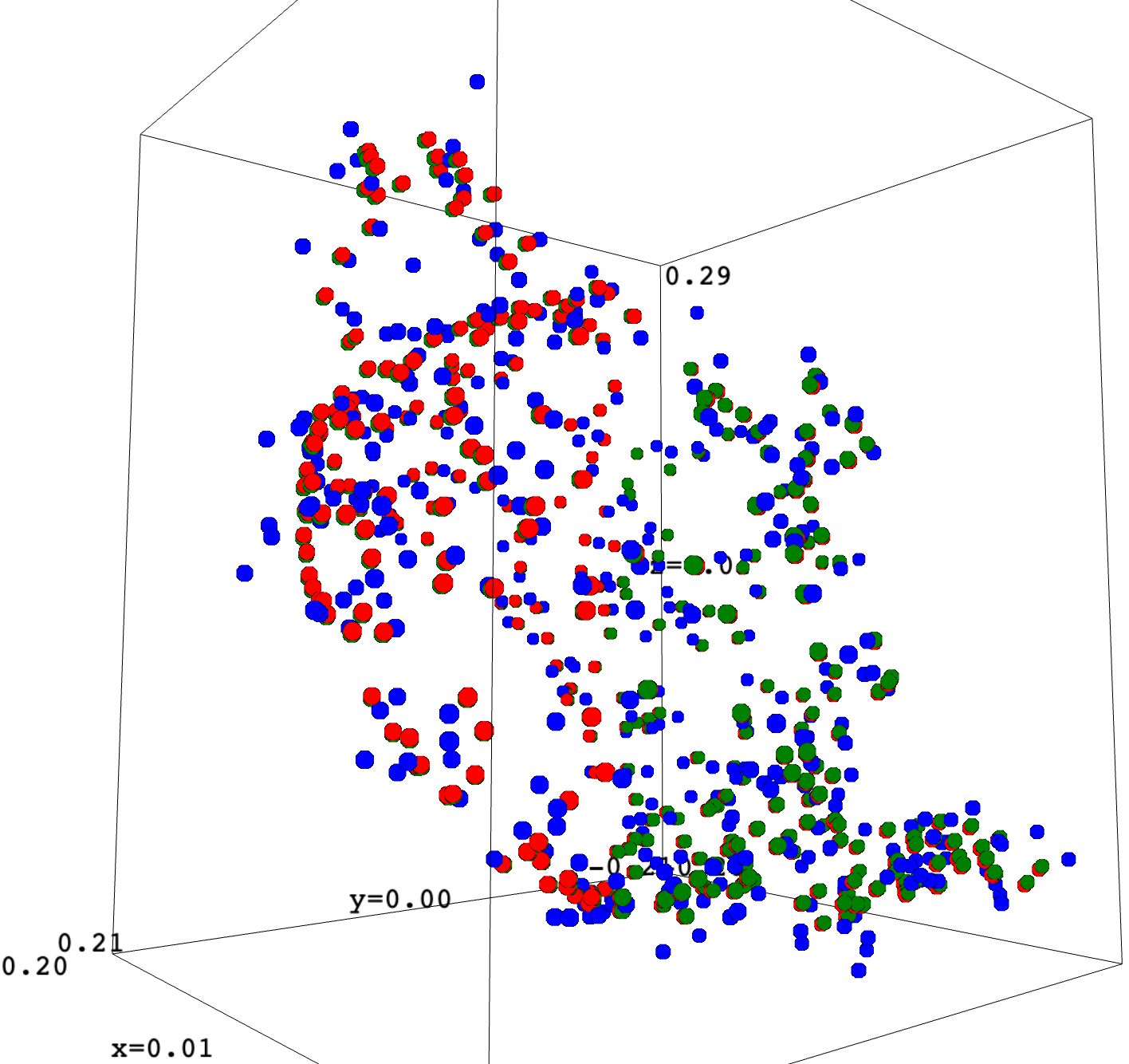}
\caption{Multiplicative noise with $\nu = 0.087$ ($\sigma = 0.1$) on the ``teapot'' cloud. Shown in the plot is the specimen cloud $Q$ (red), the noisy cloud $Q'$ (blue), and the recovered cloud $Q_{icp}$ (green). There is a significant overlap between the point clouds. }\label{teapot-mul-noise}
\end{figure}

We also produce the plots for $\delta_{spec}$ (Fig.~\ref{delta-spec-mul-noise}) and $\delta_o$ (Fig.~\ref{delta-ortho-mul-noise}) depending on $\nu$. In Fig.~\ref{teapot-mul-noise} the case of a noisy ``teapot'' cloud is shown. 

\subsection{Additive noise}

$100$ tests as described above were performed on each of the three unoccluded Caerbannog clouds \cite{caerbannog:2020}: the teapot, the bunny, and the cow. The procedure and measurements were similar to the case of multiplicative noise. 

\begin{figure}[h]
\centering
\includegraphics[width=0.33\textwidth]{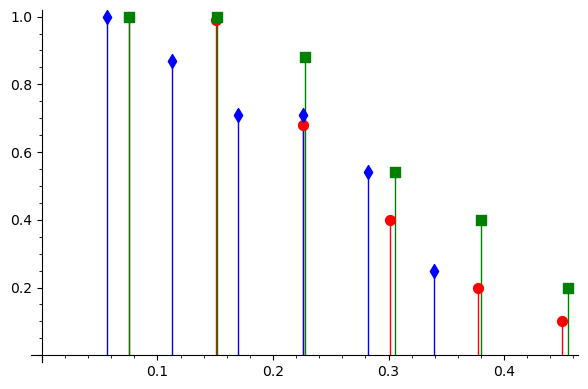}
\caption{Success rate
depending on the normalized additive noise measured over $100$ tests for each of the ``teapot'' (red circle), ``bunny'' (green square) and ``cow'' (blue lozenge) point clouds.    $\nu$ and  $\tau$ are computed for Gaussian multiplicative noise $\mathcal{N}(0, \sigma^2)$,  $\sigma \in (0.01, 0.02, \ldots, 0.06$),}
\label{success-rate-add-noise}
\end{figure}

In Fig.~\ref{success-rate-add-noise} we plot the success rates measured for different levels of noise by running $100$ tests in each case, for each of the Caerbannog clouds.  

\begin{figure}[h]
\centering
\includegraphics[width=0.36\textwidth]{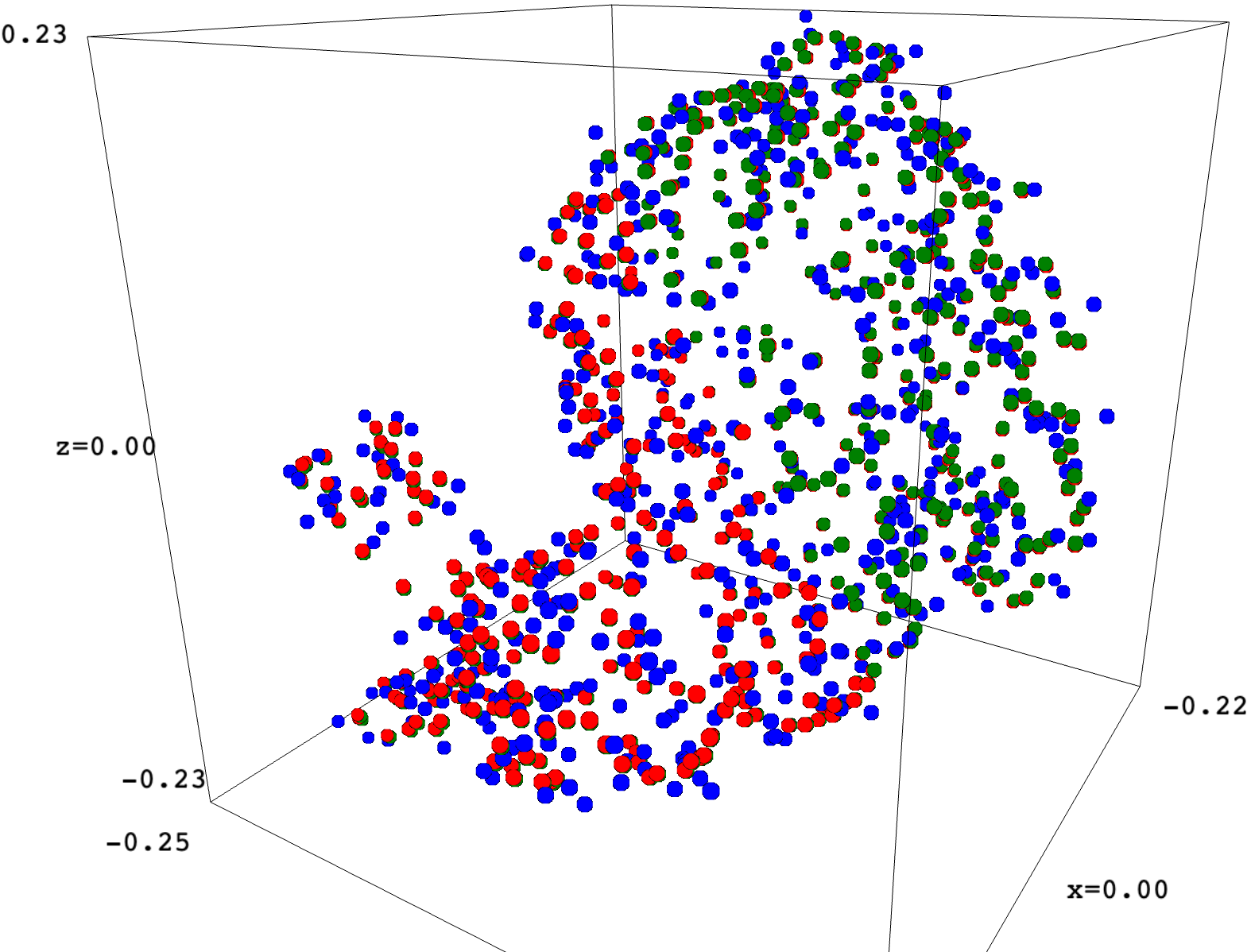}
\caption{Additive noise with $\nu = 0.074$ ($\sigma = 0.01$) on the ``bunny'' cloud. Shown in the plot is the specimen cloud $Q$ (red), the noisy cloud $Q'$ (blue), and the recovered cloud $Q_{icp}$ (green). There is a significant overlap between the point clouds. }
\label{bunny-add-noise}
\end{figure}

In Fig.~\ref{bunny-add-noise} the case of a noisy ``bunny'' cloud is shown. 

Additional plots and test statistics are available in the GitHub repository \cite{github:2022}. 

\subsection{Occluded images}

$100$ tests as described above were performed on each of the three unoccluded Caerbannog clouds \cite{caerbannog:2020}: the teapot, the bunny, and the cow. We used $P' = P$ unoccluded as a specimen cloud, and $Q' = Q \cup \Delta Q$ as its occluded image. Here $Q = O\, P\, S$, as described in Section~\ref{occlusion}, and $\Delta Q$ was created as follows. First, we determine the rectangular bounding box $B$ for $Q$ with sides parallel to the coordinate planes and then generate uniformly inside $B$  random points $\Delta Q$. The cardinality of $\Delta Q$ is controlled by the level of occlusion $\alpha$: namely $|\Delta Q|$ is the integer part of $ \alpha \, |Q|$. 

In our tests, we used $\alpha \in \{0.2, 0.4, \cdots, 1.2\}$.

\begin{comment}

and measured the following values:

\begin{itemize}
    \item the added occlusion (normalized): $\nu = \| Q' - Q \|_2 / \|P\|_2$;
    \item the normalized distance to the occluded image $Q'$: \\$\delta = \| Q' - Q_{icp} \|_2 / \|P\|_2$;
    \item the normalized distance to the actual specimen image: $\delta_{spec} = \| Q - U_{icp}\,P\,S \|_2 / \|P\|_2$; 
    \item the success rate of the test batch: $100$ tests with success criterion $\delta_{spec} \leq 0.05$;
    \item the distance to the original orthogonal transformation \\$\delta_o = \| U_{icp} - O \|_2$;
    \item the ICP change in the (normalized) distance: \\ $\delta^{icp} = \left( \| Q' - Q_{init} \|_2 - \| Q' - Q_{icp} \|_2 \right) / \|  P \|_2$;
    \item the ICP change to the orthogonal transformation: \\ $\delta^{icp}_o = \| U_{init} - U_{icp} \|_2$.
\end{itemize}
\end{comment}

\begin{figure}[h]
\centering
\includegraphics[width=0.33\textwidth]{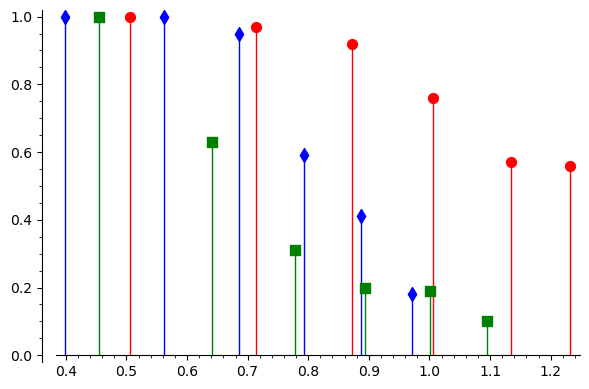}
\caption{Success rate  depending on the normalized occlusion  averaged over $100$ tests for each of the ``teapot'' (red circle), ``bunny'' (green square) and ``cow'' (blue lozenge) point clouds. $\nu$ and $\tau$ are computed for occlusion levels  $\alpha \in (0.2, 0.4, \ldots, 1.2)$.}\label{success-rate-occluded}

\vspace{0.2in}

\centering
\includegraphics[width=0.33\textwidth]{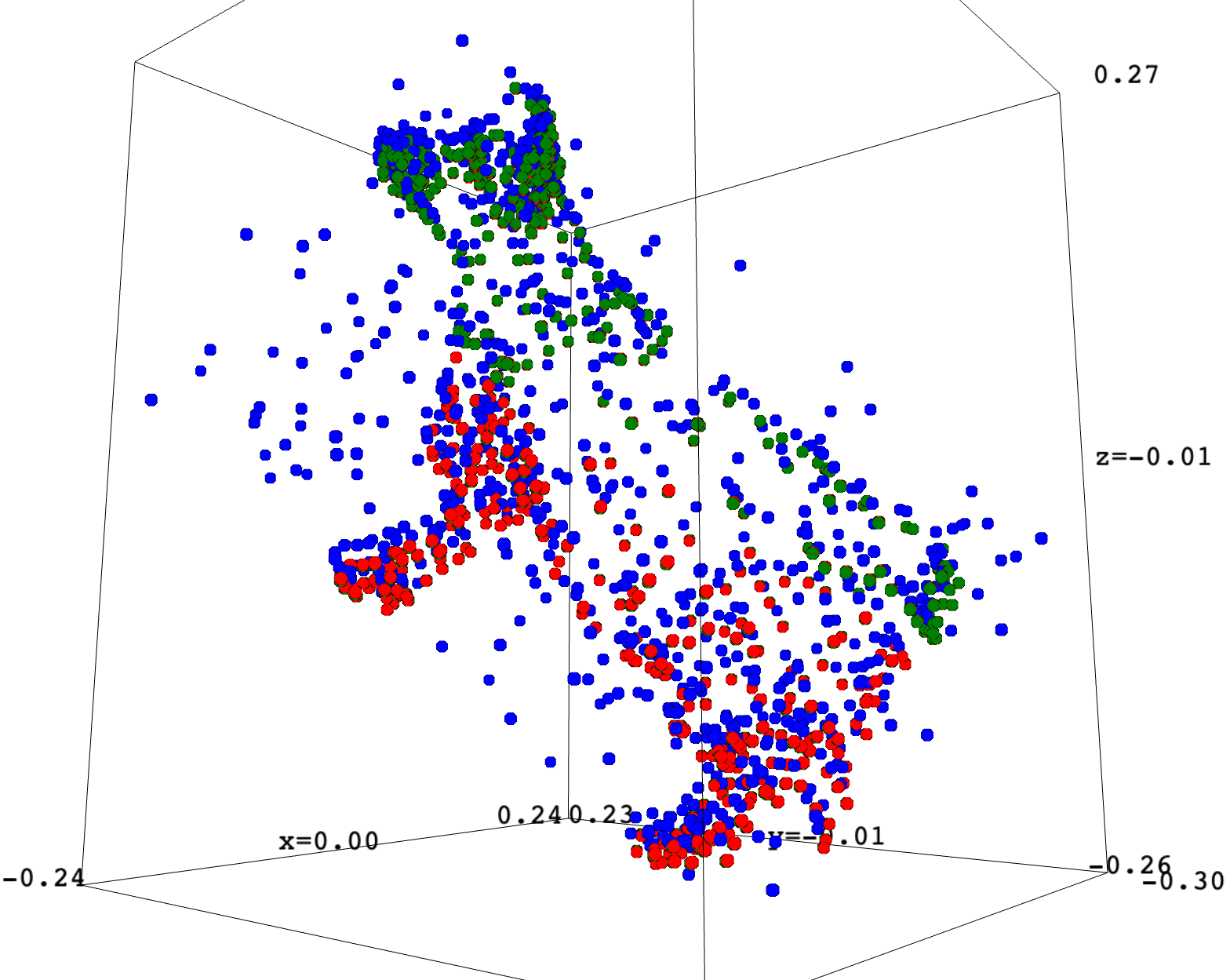}
\caption{Occlusion with $\nu = 0.57$ ($\alpha = 0.4$) on the ``cow'' cloud. Shown  is the specimen cloud $Q$ (red), the occluded cloud $Q'$ (blue), and the recovered cloud $Q_{icp}$ (green). There is a significant overlap between the point clouds.}\label{cow-occluded}
\end{figure}

In Fig.~\ref{success-rate-occluded} we plot the success rates measured for different levels of noise by running $100$ tests in each case, for each of the Caerbannog clouds. 

In Fig.~\ref{cow-occluded} the case of a noisy ``cow'' cloud is shown.  Additional plots are available in the GitHub repository \cite{github:2022}.

\subsection{Superposition of errors}
\label{super}

\begin{figure}[h]
\centering
\includegraphics[width=0.45\textwidth]{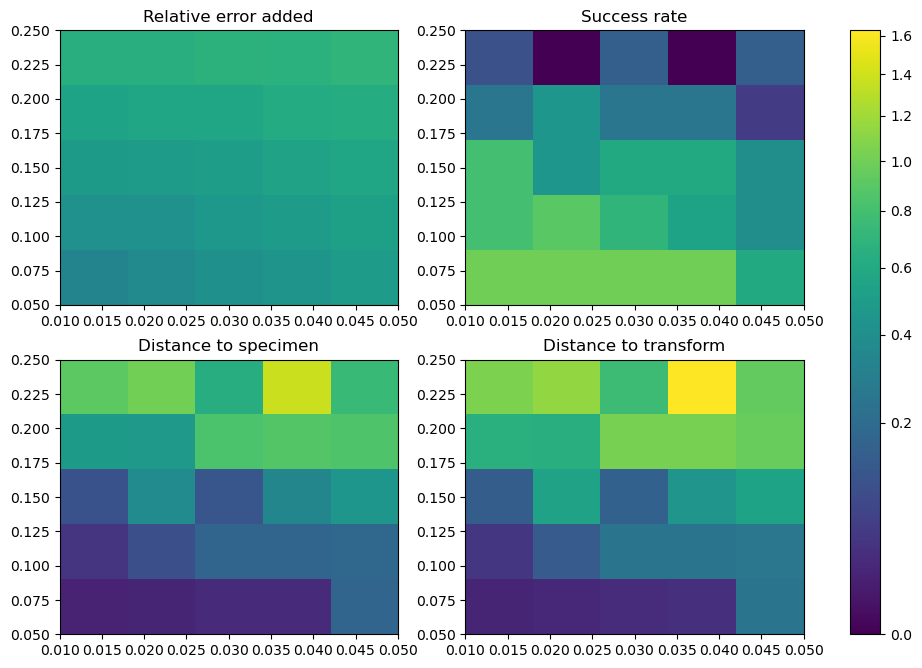}
\caption{Colormap for the statistics of tests on the ``teapot'' point cloud matching  $P$ to the occluded $Q'$ with additive and multiplicative noises superimposed. Occlusion is formed by adding uniformly distributed points in the point cloud $Q$ bounding box: occlusion points are $5\%$ of the original cloud cardinality. The noise parameters: $\sigma$ for $\mathcal{N}(0, \sigma)$ additive noise is indicated on the horizontal axis,  $\sigma$ for $\mathcal{N}(1, \sigma)$ multiplicative noise is indicated on the vertical axis.}\label{teapot_occ_005}

\vspace{0.2in}

\centering
\includegraphics[width=0.45\textwidth]{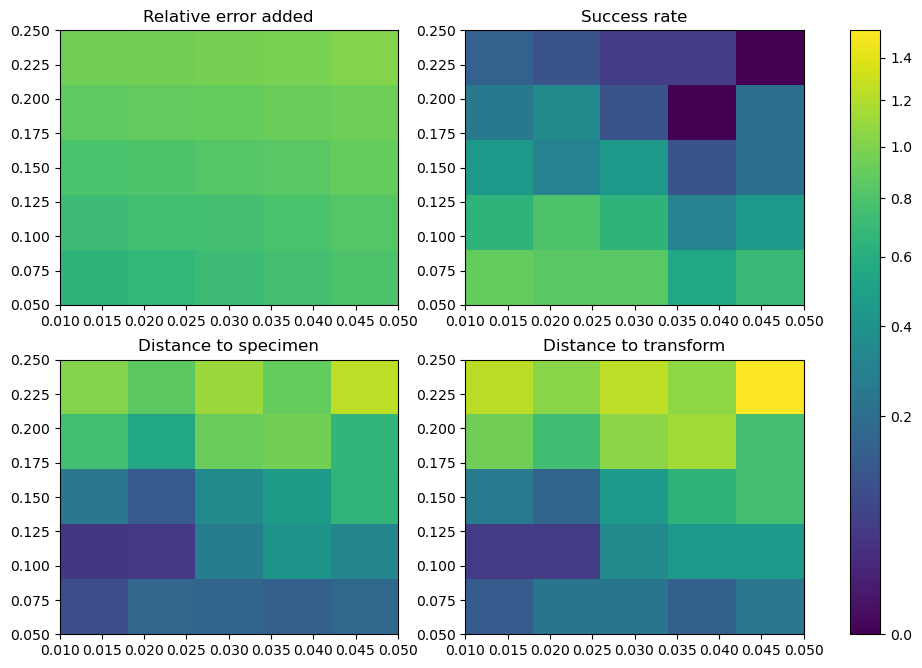}
\caption{Colormap for the statistics of tests on the ``teapot'' point cloud matching $P$ to the occluded $Q'$ with additive and multiplicative noises superimposed. Occlusion is formed by adding uniformly distributed points in the point cloud $Q$ bounding box: occlusion points are $25\%$ of the original cloud cardinality. The noise parameters: $\sigma$ for $\mathcal{N}(0, \sigma)$ additive noise is indicated on the horizontal axis,  $\sigma$ for $\mathcal{N}(1, \sigma)$ multiplicative noise is indicated on the vertical axis.}\label{teapot_occ_025}
\end{figure}

We provide the results of $100$ tests performed for various levels of occlusion, additive, and multiplicative noises superimposed on each of the Caerbannog clouds. For example, in Figures~\ref{teapot_occ_005} -- \ref{teapot_occ_025} the case of the ``teapot'' cloud is presented. In each figure, the level of occlusion is fixed, while we provide the dependence of the measured parameters on the values of $\sigma$ for additive and multiplicative noises. 
\begin{comment}
    Namely, we measure the total error added $\nu$, the distance to specimen $\delta_{spec}$, the distance $\delta_o$ to the original orthogonal transformation, and the success rate $\tau$ (a test is declared successful if $\delta_{spec} \leq 0.05$). 
\end{comment}
Additional plots are available on GitHub \cite{github:2022}.

\subsection{Registering multiple scans}

\begin{figure}[h]

\centering
{
~\hspace{0.1in}~
%\begin{minipage}{0.2\textwidth}
\includegraphics[height=0.18\textheight]{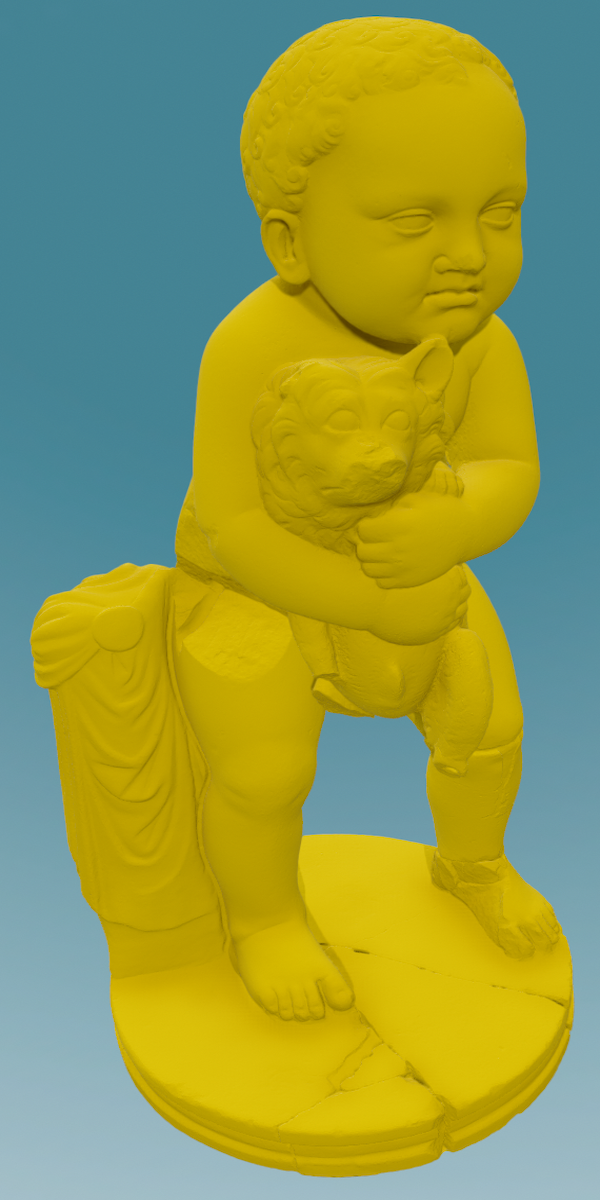}
%\end{minipage}~\hspace{-0.8in}~
%\begin{minipage}{0.2\textwidth}
\includegraphics[height=0.18\textheight]{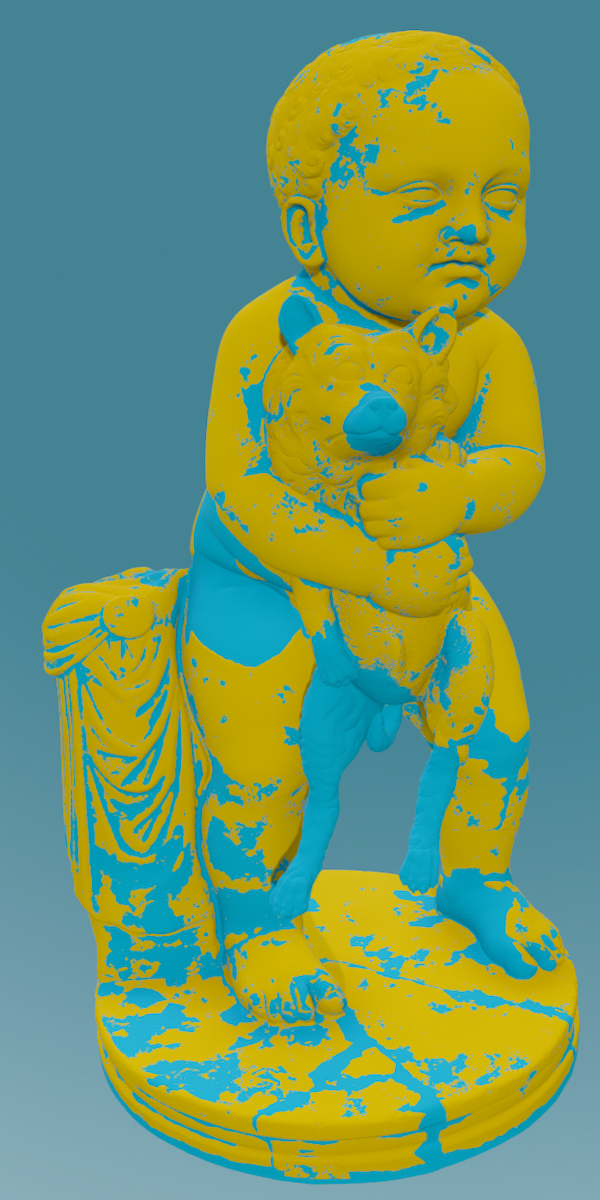}
%\end{minipage}~\hspace{-0.8in}~
%\begin{minipage}{0.25\textwidth}
\includegraphics[height=0.18\textheight]{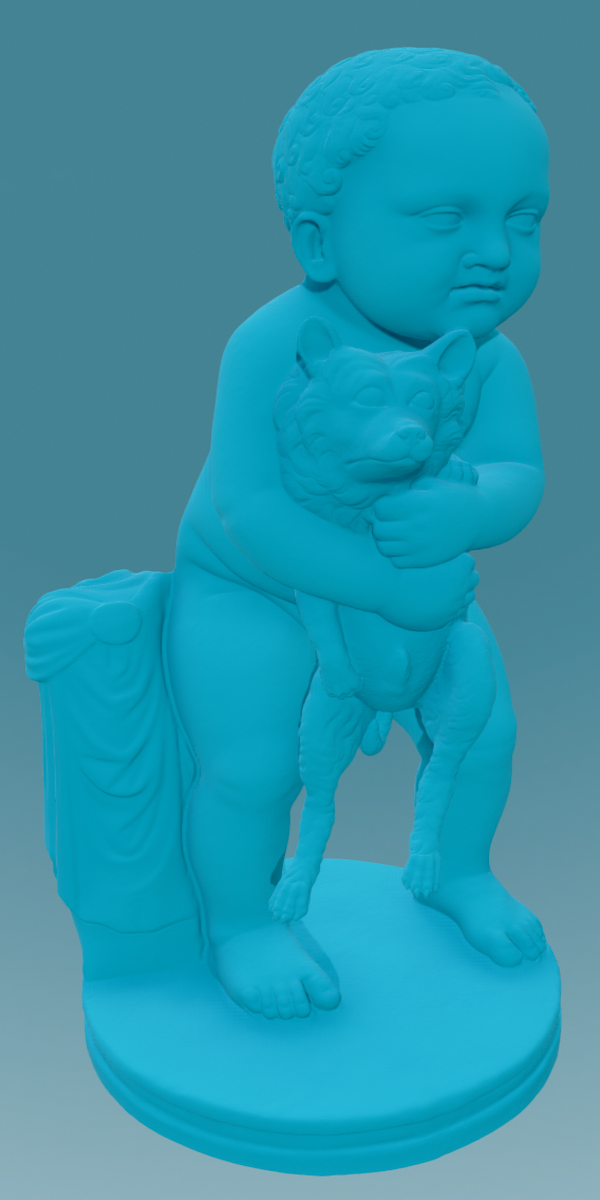}
%\end{minipage}
}
\caption{``Enfant au Chien'' before (left) and after (right) restoration (available from \cite{3-d-scans}). The ground truth of matching the sculptures (center).}\label{statue-broken-restored-0}

\end{figure}

\begin{figure}[h]

\centering
{
~\hspace{0.4in}~
%\begin{minipage}{0.35\textwidth}
\includegraphics[height=0.18\textheight]{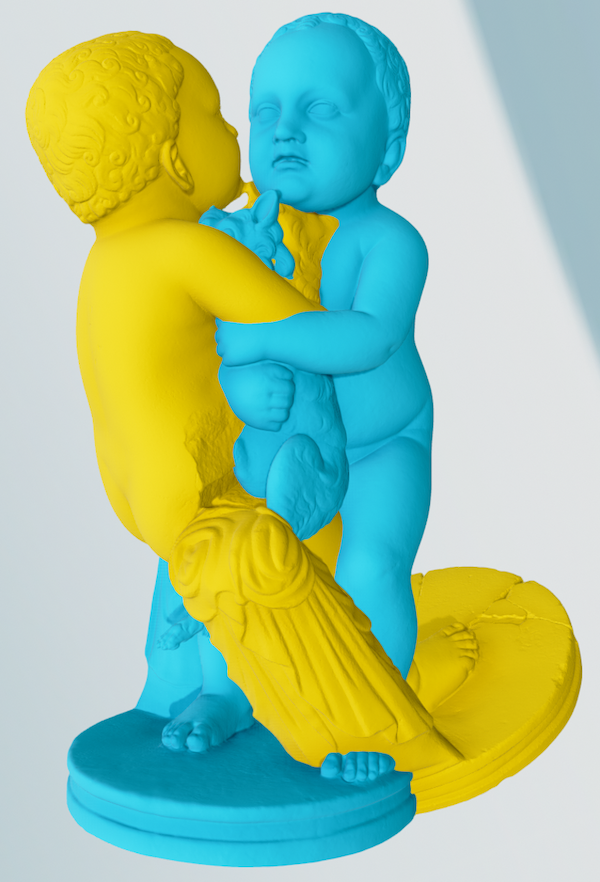}
%\end{minipage}~\hspace{-1.05in}~
%\begin{minipage}{0.35\textwidth}
\includegraphics[height=0.18\textheight]{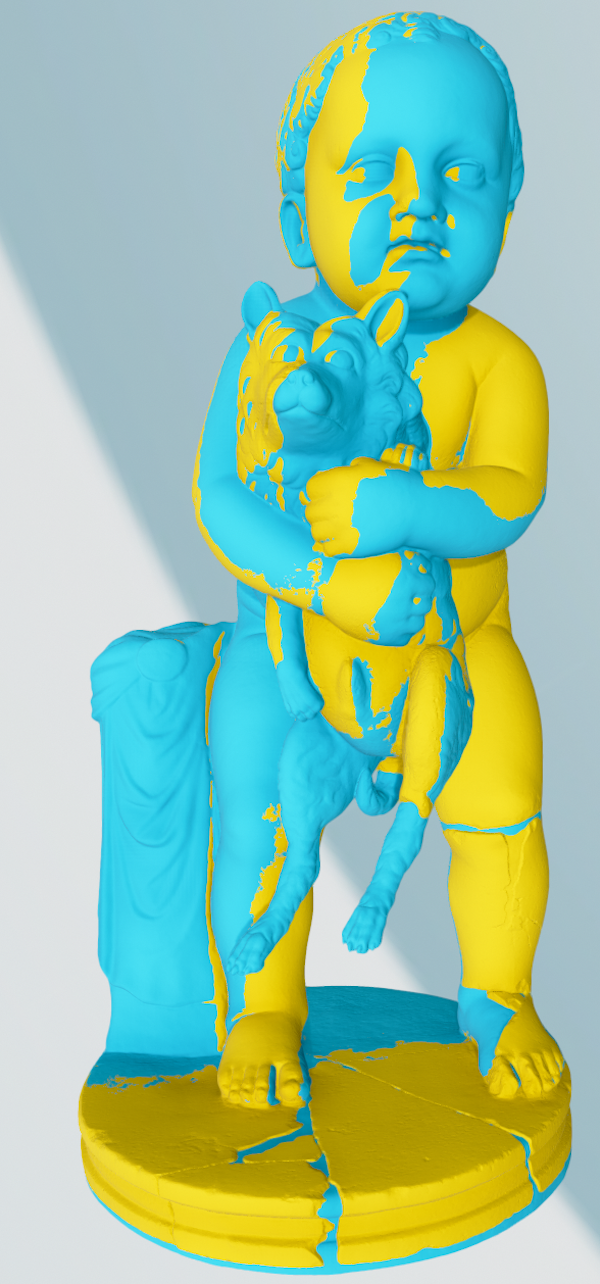}
%\end{minipage}
}
\caption{``Enfant au Chien'' ICP registration: without initialization (left) and with initialization (right).}\label{statue-init-vs-no-init-0}

\end{figure}

\begin{figure}[h]

\centering
{
~\hspace{0.05in}~
%\begin{minipage}{0.25\textwidth}
\includegraphics[height=0.18\textheight]{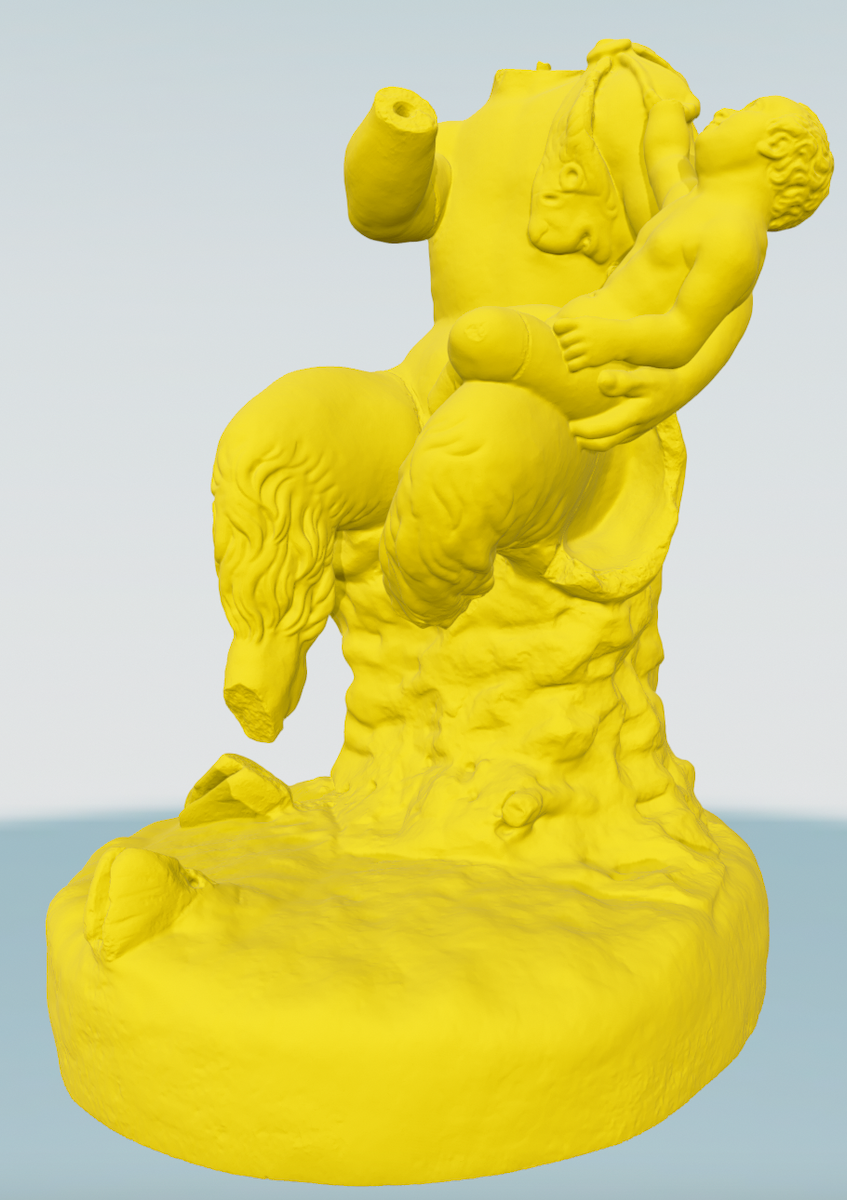}
%\end{minipage}~\hspace{-0.65in}~
%\begin{minipage}{0.25\textwidth}
\includegraphics[height=0.18\textheight]{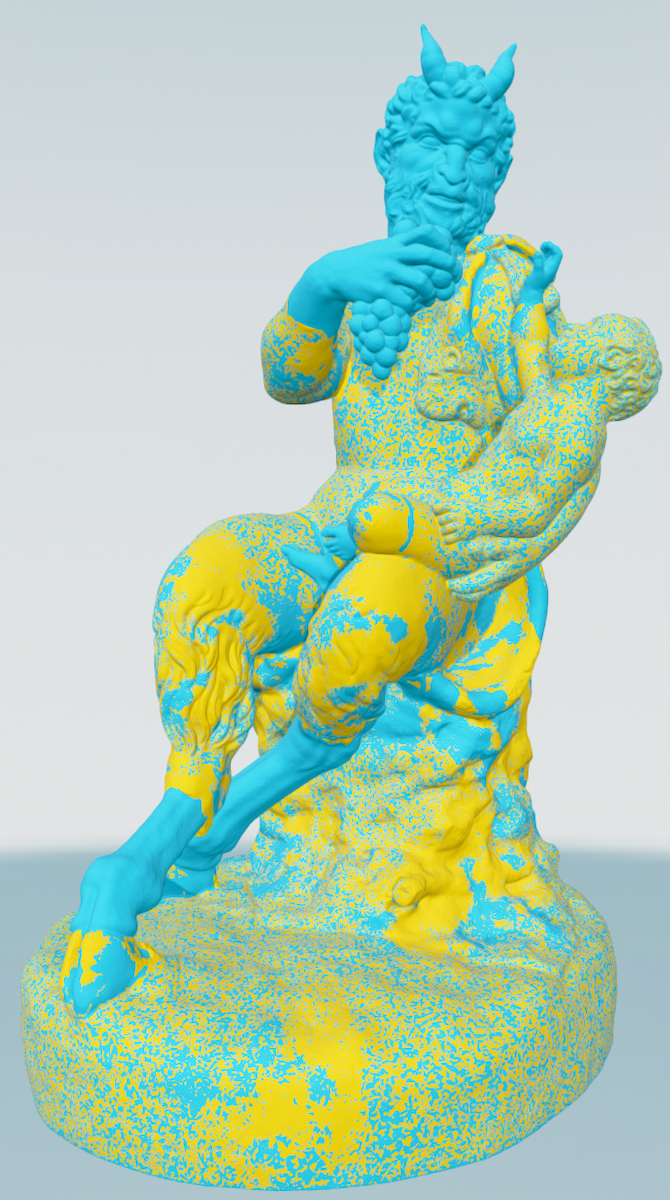}
%\end{minipage}~\hspace{-0.90in}~
%\begin{minipage}{0.25\textwidth}
\includegraphics[height=0.18\textheight]{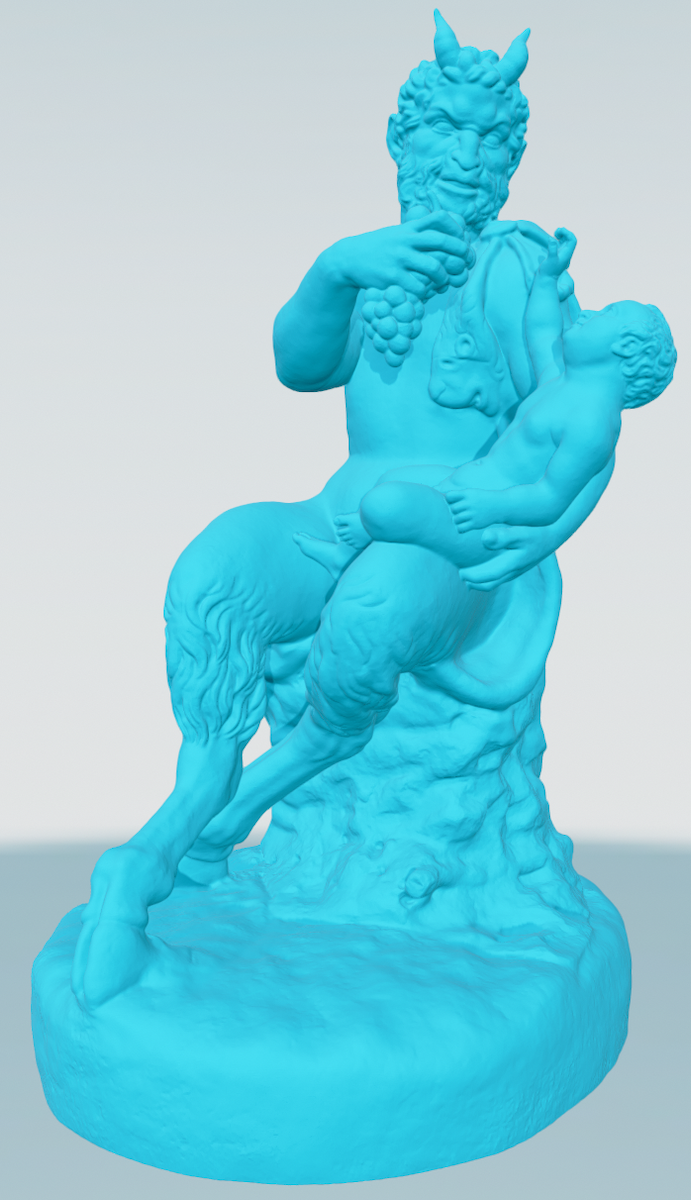}
%\end{minipage}
}
\caption{``Pan and Bacchus'' before (left) and after (right) restoration (available from \cite{3-d-scans}). The ground truth of matching the sculptures (center).}\label{statue-broken-restored-1}

\end{figure}

\begin{figure}[h]

\centering
{
~\hspace{0.05in}~
%\begin{minipage}{0.35\textwidth}
\includegraphics[height=0.18\textheight]{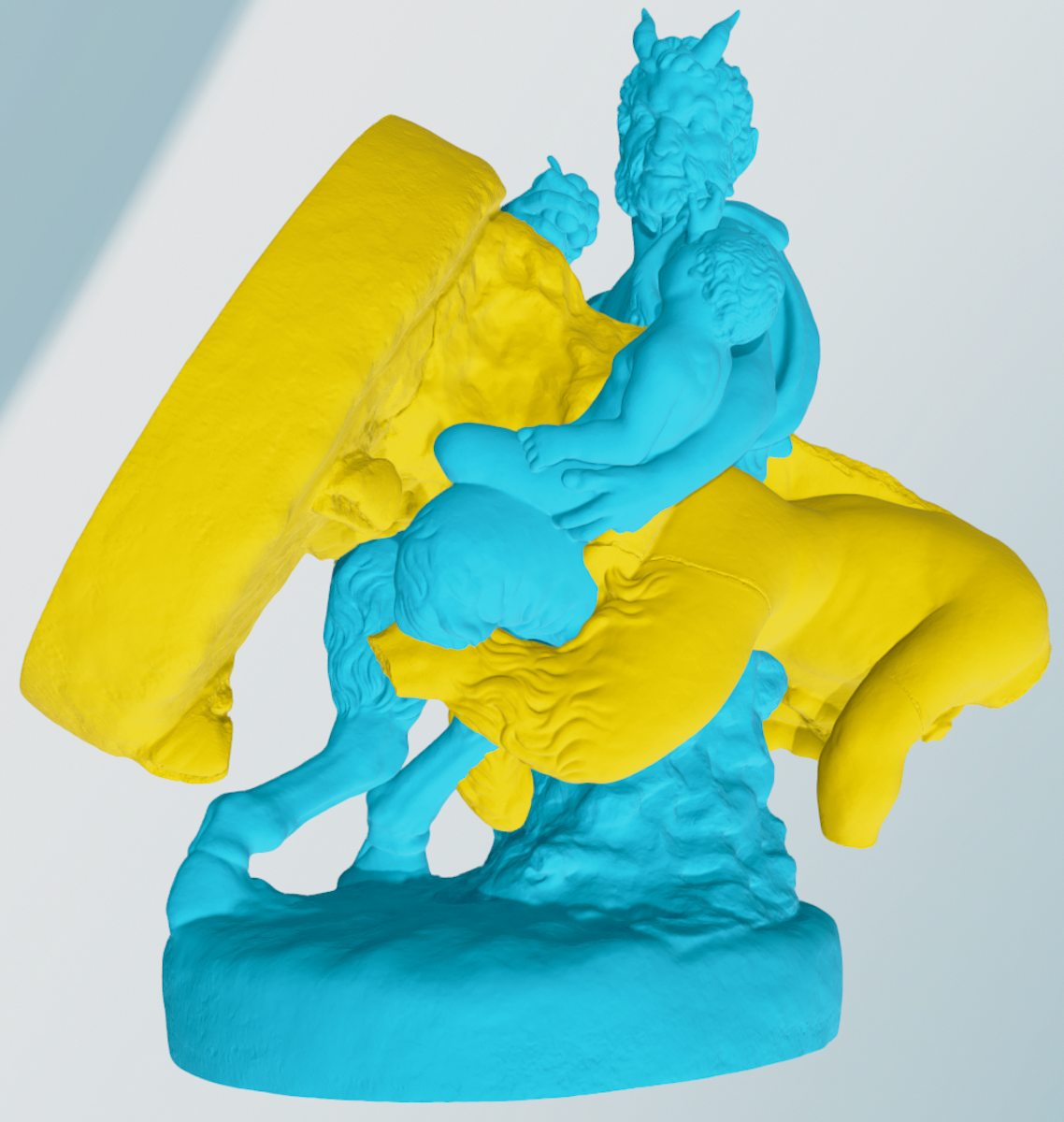}
%\end{minipage}~\hspace{-0.43in}~
%\begin{minipage}{0.35\textwidth}
\includegraphics[height=0.18\textheight]{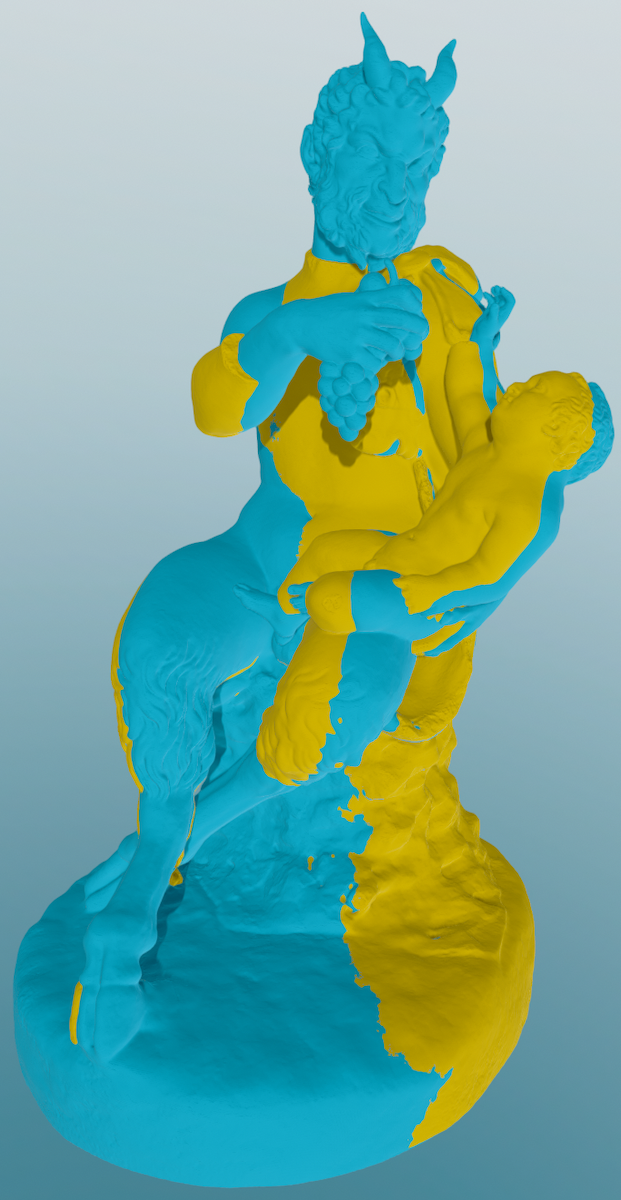}
%\end{minipage}
}
\caption{``Pan and Bacchus'' ICP registration: without initialization (left) and with initialization (right).}\label{statue-init-vs-no-init-1}

\end{figure}

\begin{figure}[h]

\centering
{
~\hspace{0.05in}~
%\begin{minipage}{0.25\textwidth}
\includegraphics[height=0.18\textheight]{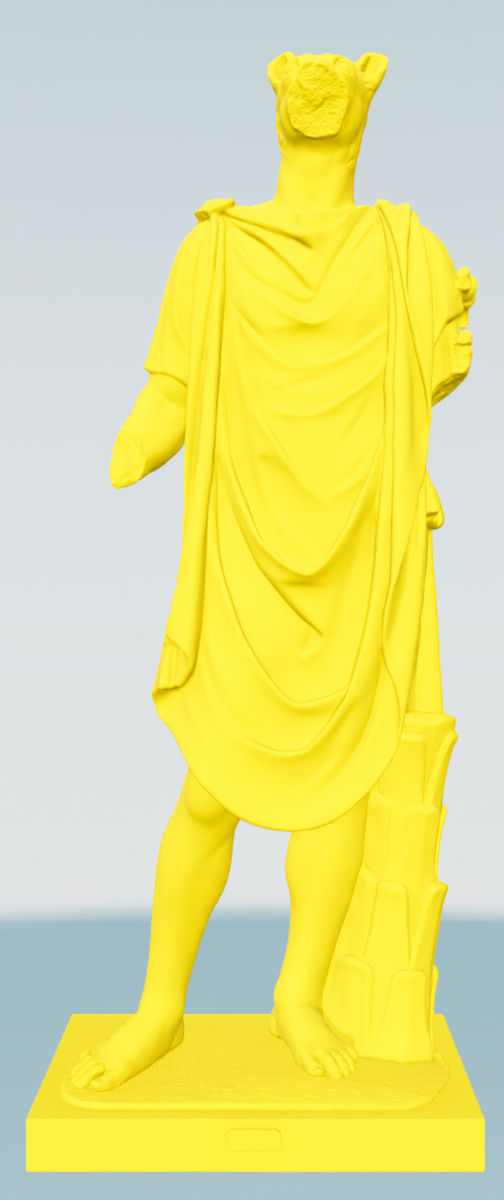}
%\end{minipage}~\hspace{-0.89in}~
%\begin{minipage}{0.25\textwidth}
\includegraphics[height=0.18\textheight]{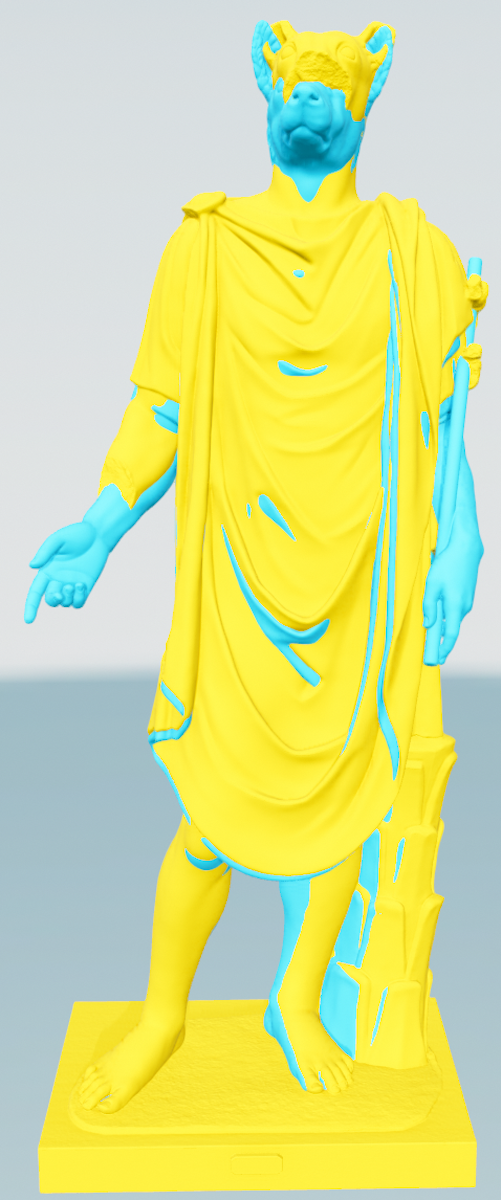}
%\end{minipage}~\hspace{-0.90in}~
%\begin{minipage}{0.25\textwidth}
\includegraphics[height=0.18\textheight]{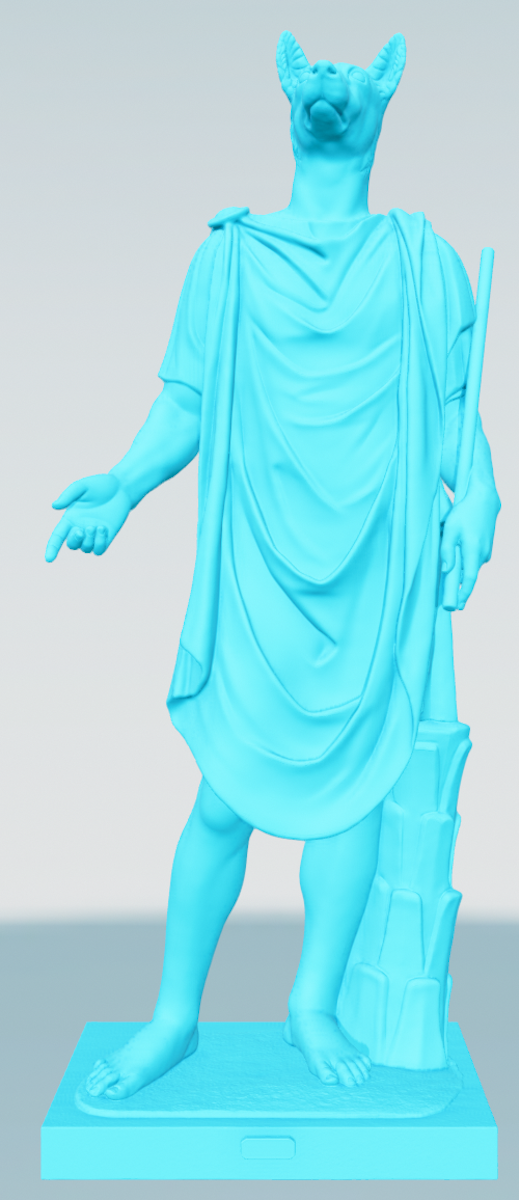}
%\end{minipage}
}
\caption{``Hermanibus'' before (left) and after (right) restoration (available from \cite{3-d-scans}). The ground truth of matching the sculptures (center).}\label{statue-broken-restored-2}

\end{figure}

\begin{figure}[h]

\centering
{
~\hspace{0.05in}~
%\begin{minipage}{0.35\textwidth}
\includegraphics[height=0.18\textheight]{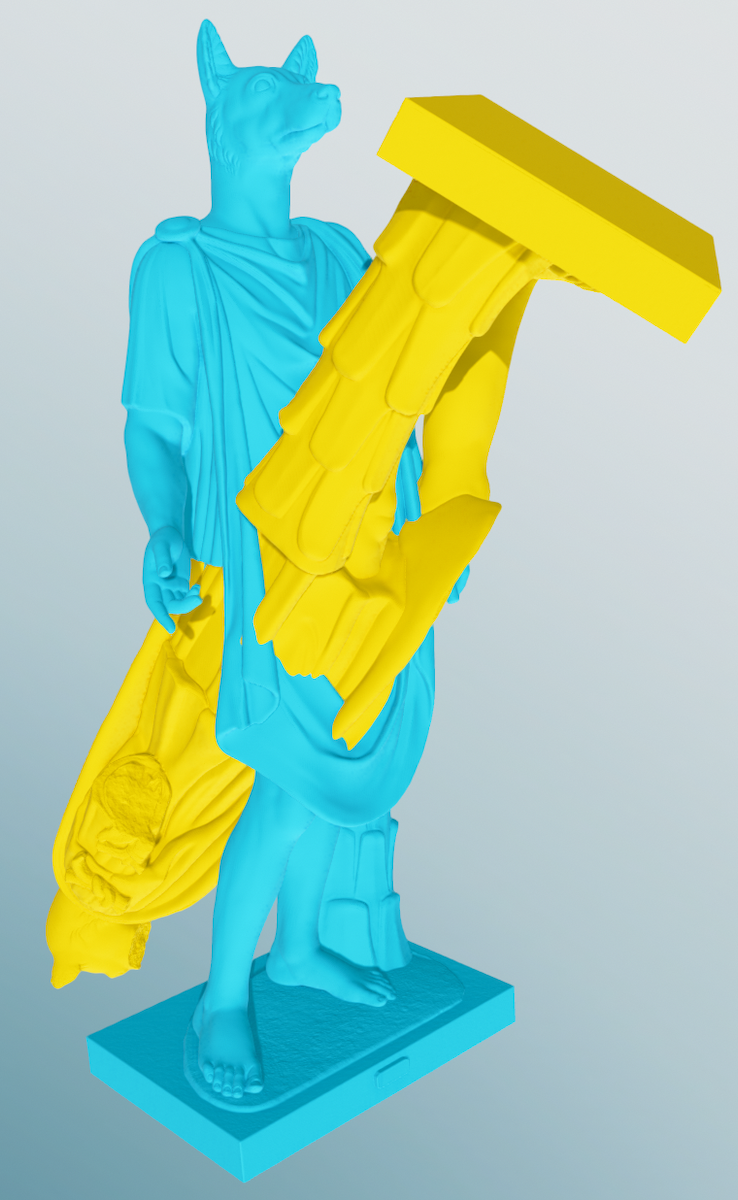}
%\end{minipage}~\hspace{-0.90in}~
%\begin{minipage}{0.35\textwidth}
\includegraphics[height=0.18\textheight]{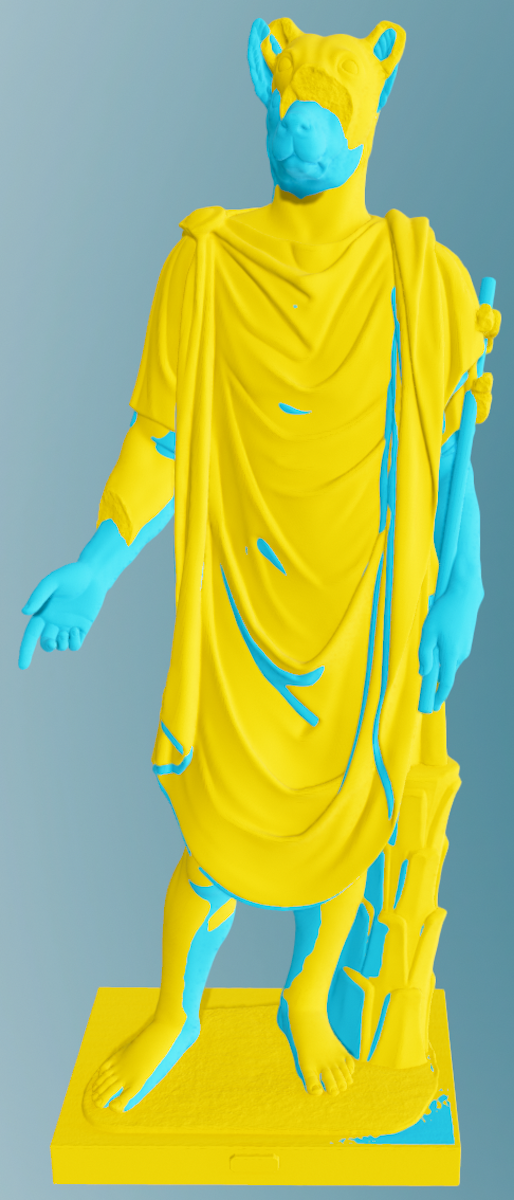}
%\end{minipage}
}
\caption{``Hermanibus'' ICP registration: without initialization (left) and with initialization (right).}\label{statue-init-vs-no-init-2}

\end{figure}

We compare two scanned images of the same sculpture before and after restoration in several cases shown in Figures~\ref{statue-broken-restored-0}, \ref{statue-broken-restored-1}, \ref{statue-broken-restored-2} (see Section~\ref{appendix}, also for Figures~\ref{statue-init-vs-no-init-0}, \ref{statue-init-vs-no-init-1}, \ref{statue-init-vs-no-init-2}).
 
Our {\it E--Init} algorithm performs well in this case, as shown in Figures~\ref{statue-init-vs-no-init-0}~(right), \ref{statue-init-vs-no-init-1}~(right), \ref{statue-init-vs-no-init-2}~(right), used in conjunction with the out--of--the--box ICP supplied in \texttt{open3d} \cite{open3d}. The latter applied without {\it E--Init} fails to register properly, as shown in Figures~\ref{statue-init-vs-no-init-0}~(left), \ref{statue-init-vs-no-init-1}~(left), \ref{statue-init-vs-no-init-2}~(left). We used the point--to--plane version of ICP in all instances. 

The ground truth for matching the broken and restored sculptures is shown in Figures~\ref{statue-broken-restored-0}~(center), \ref{statue-broken-restored-1}~(center), \ref{statue-broken-restored-2}~(center). It is worth mentioning that the statue has relatively light damages only in Figure~\ref{statue-broken-restored-0}, while in Figure~\ref{statue-broken-restored-1} the broken and restored statues differ significantly. Also, in Figure~\ref{statue-broken-restored-2} the ground truth alignment shows that the broken and restored images do not fully match. 

The code and mesh data used  are available on GitHub \cite{github:2022}.

\subsection{Comparison to other algorithms}

We use GO--ICP \cite{yang2015go} as a state--of--the--art reference algorithm as it provides provably convergent branch--and--bound strategy for global search over the orthogonal group $\mathrm{O}(3)$. In our experiments, $10$ tests were performed on each of the Caerbannog point clouds \cite{caerbannog:2020} for comparison. With multiplicative noise $\mathcal{N}(1, \sigma)$ having $\sigma=0.1$ 

 It is worth mentioning that normal multiplicative noise with $\sigma = 0.1$ (in which case $0.08 \leq \nu \leq 0.093$) requires the MSE parameter of GO--ICP to be set to $0.0001$. Lower values of the MSE parameter (e.g. $0.001$) result in the branch--and--bound method choosing the wrong branch at times. The time needed to find the right parameters,
which are necessary to run GO--ICP, were not taken into consideration.
The results showing a much larger computation time for GO--ICP than E--init are presented in Table~\ref{table}.

\begin{table}[h!]
\centering
\begin{tabular}{||l c c c||} 
 \hline
 Model & Average time(s) & $\delta_{spec}$  & $\delta_{o}$  \\ [0.5ex] 
 \hline\hline
  Teapot GO--ICP & 75.90 & 0.008 & 0.009 \\
  Teapot E--init & 2.77 & 0.006 & 0.007 \\ \hline
  Bunny  GO--ICP &103.64& 0.011&0.012\\
  Bunny E--init & 6.35&0.004&0.005\\ \hline
  Cow GO--ICP &18.17&0.011&0.012\\
  Cow  E--init &8.34&0.005&0.006\\
  
\hline
\end{tabular}
\caption{Comparison of E--init with GO--ICP.}\label{table}
\end{table}

\subsection{GitHub repository}

The SageMath code used to perform the numerical experiments is available on GitHub at \href{https://github.com/sashakolpakov/icp-init}{https://github.com/sashakolpakov/icp--init}. 
The open source computer algebra system SageMath (freely available at \href{https://www.sagemath.org/}{https://www.sagemath.org/}) is required to run the code.

\section{Conclusion}
This note presents  an approach to initialize the Iterative Closest Point (ICP) algorithm  to match unlabelled point clouds related by rigid transformations so that it finds good solutions. The method is especially pertinent when there is a large overlap between the centered point clouds and is based on matching the ellipsoids defined by the points' covariance matrices and then trying the various principal half--axes matchings that differ by elements of the finite reflection group of the ellipse's axial planes. We derive bounds on the robustness of our approach to noise and  experiments  confirm our theoretical findings and the usefulness of our initialization. 

% insert where needed to balance the two columns on the last page with
% biographies
%\newpage

% You can push biographies down or up by placing
% a \vfill before or after them. The appropriate
% use of \vfill depends on what kind of text is
% on the last page and whether or not the columns
% are being equalized.

%\vfill

% Can be used to pull up biographies so that the bottom of the last one
% is flush with the other column.
%\enlargethispage{-5in}

% that's all folks

\noindent

\bibliographystyle{IEEEtran}
\bibliography{IEEEabrv,refs}

% Generated by IEEEtran.bst, version: 1.14 (2015/08/26)
\begin{thebibliography}{10}
\providecommand{\url}[1]{#1}
\csname url@samestyle\endcsname
\providecommand{\newblock}{\relax}
\providecommand{\bibinfo}[2]{#2}
\providecommand{\BIBentrySTDinterwordspacing}{\spaceskip=0pt\relax}
\providecommand{\BIBentryALTinterwordstretchfactor}{4}
\providecommand{\BIBentryALTinterwordspacing}{\spaceskip=\fontdimen2\font plus
\BIBentryALTinterwordstretchfactor\fontdimen3\font minus
  \fontdimen4\font\relax}
\providecommand{\BIBforeignlanguage}[2]{{%
\expandafter\ifx\csname l@#1\endcsname\relax
\typeout{** WARNING: IEEEtran.bst: No hyphenation pattern has been}%
\typeout{** loaded for the language `#1'. Using the pattern for}%
\typeout{** the default language instead.}%
\else
\language=\csname l@#1\endcsname
\fi
#2}}
\providecommand{\BIBdecl}{\relax}
\BIBdecl

\bibitem{app}
\BIBentryALTinterwordspacing
H.~Si, J.~Qiu, and Y.~Li, ``A review of point cloud registration algorithms for
  laser scanners: Applications in large--scale aircraft measurement,''
  \emph{Applied Sciences}, vol.~12, no.~20, 2022. [Online]. Available:
  \url{https://www.mdpi.com/2076-3417/12/20/10247}
\BIBentrySTDinterwordspacing

\bibitem{tut}
L.~Li, R.~Wang, and X.~Zhang, ``A tutorial review on point cloud registrations:
  Principle, classification, comparison, and technology challenges,''
  \emph{Mathematical Problems in Engineering}, 2021.

\bibitem{121791}
P.~Besl and N.~D. McKay, ``A method for registration of 3--{D} shapes,''
  \emph{IEEE Transactions on Pattern Analysis and Machine Intelligence},
  vol.~14, no.~2, pp. 239--256, 1992.

\bibitem{132043}
Y.~Chen and G.~Medioni, ``Object modeling by registration of multiple range
  images,'' in \emph{Proceedings. 1991 IEEE International Conference on
  Robotics and Automation}, 1991, pp. 2724--2729 vol.3.

\bibitem{enwiki:1095051621}
\BIBentryALTinterwordspacing
{Wikipedia contributors}, ``Iterative closest point --- {Wikipedia}{,} the free
  encyclopedia,'' 2022. [Online]. Available:
  \url{https://en.wikipedia.org/w/index.php?title=Iterative_closest_point&oldid=1095051621}
\BIBentrySTDinterwordspacing

\bibitem{Kabsch:a12999}
\BIBentryALTinterwordspacing
W.~Kabsch, ``A solution for the best rotation to relate two sets of vectors,''
  \emph{Acta Crystallographica Section A}, vol.~32, no.~5, pp. 922--923, 1976.
  [Online]. Available: \url{https://doi.org/10.1107/S0567739476001873}
\BIBentrySTDinterwordspacing

\bibitem{yang2015go}
J.~Yang, H.~Li, D.~Campbell, and Y.~Jia, ``{GO--ICP}: {A} globally optimal
  solution to 3{D} {ICP} point--set registration,'' \emph{IEEE transactions on
  pattern analysis and machine intelligence}, vol.~38, no.~11, pp. 2241--2254,
  2015.

\bibitem{huang2021comprehensive}
X.~Huang, G.~Mei, J.~Zhang, and R.~Abbas, ``A comprehensive survey on point
  cloud registration,'' \emph{arXiv preprint arXiv:2103.02690}, 2021.

\bibitem{serafin2015nicp}
J.~Serafin and G.~Grisetti, ``Nicp: Dense normal based point cloud
  registration,'' in \emph{2015 IEEE/RSJ International Conference on
  Intelligent Robots and Systems (IROS)}.\hskip 1em plus 0.5em minus
  0.4em\relax IEEE, 2015, pp. 742--749.

\bibitem{SERAFIN201791}
------, ``Using extended measurements and scene merging for efficient and
  robust point cloud registration,'' \emph{Robotics and Autonomous Systems},
  vol.~92, pp. 91--106, 2017.

\bibitem{Makovetskii}
A.~Makovetskii, S.~Voronin, V.~Kober, , and A.~Voronin, ``A regularized point
  cloud registration approach for orthogonal transformations,'' \emph{Journal
  of Global Optimization}, 2022.

\bibitem{arnold72}
V.~Arnold, ``Modes and quasimodes,'' \emph{Func. Anal. Appl.}, vol.~6, no.~2,
  pp. 94--101, 1972.

\bibitem{Ahlswede:2002}
R.~Ahlswede and A.~Winter, ``Strong converse for identification via quantum
  channels,'' \emph{IEEE Trans. Inf. Theory}, vol.~48, no.~3, pp. 569--579,
  2002, [Addendum ibid 49(1):346, 2003].

\bibitem{caerbannog:2020}
\BIBentryALTinterwordspacing
S.~Raghupathi, N.~Brunhart-Lupo, and K.~Gruchalla, ``{C}aerbannog point clouds.
  {N}ational {R}enewable {E}nergy {L}aboratory,'' 2020. [Online]. Available:
  \url{https://data.nrel.gov/submissions/153}
\BIBentrySTDinterwordspacing

\bibitem{github:2022}
\BIBentryALTinterwordspacing
A.~Kolpakov and M.~Werman, ``Sage{M}ath worksheets for {ICP} initialization,''
  2022. [Online]. Available: \url{https://github.com/sashakolpakov/icp-init}
\BIBentrySTDinterwordspacing

\bibitem{open3d}
Q.-Y. Zhou, J.~Park, and V.~Koltun, ``{Open3D}: {A} modern library for {3D}
  data processing,'' \emph{arXiv:1801.09847}, 2018.

\bibitem{3-d-scans}
\BIBentryALTinterwordspacing
O.~Laric, ``{T}hree {D} {S}cans,'' 2012. [Online]. Available:
  \url{https://threedscans.com/}
\BIBentrySTDinterwordspacing

\end{thebibliography}

\begin{IEEEbiography}[{\includegraphics[width=1in,height=1.25in,clip,keepaspectratio]{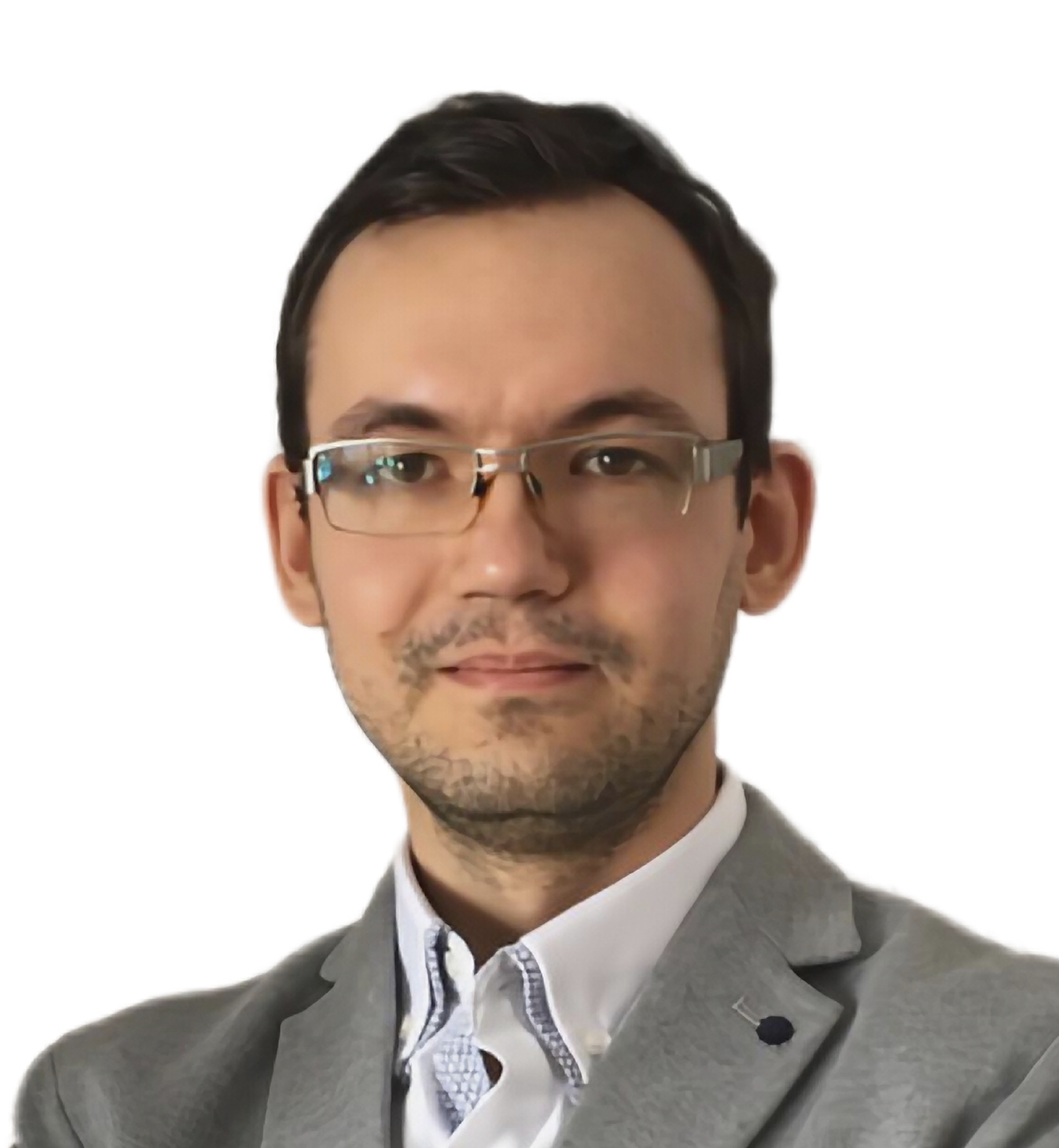}}]{Alexander Kolpakov}
Alexander Kolpakov received his Ph.D. degree from the University of Fribourg, Switzerland, in 2013. Currently, he is an Assistant Professor at the University of Neuch\^atel. His research interests are in the domains of Riemannian geometry and group theory, and also  applications of geometry to data science and image processing. 
\end{IEEEbiography}

\begin{IEEEbiography}[{\includegraphics[width=1in,height=1.25in,clip,keepaspectratio]{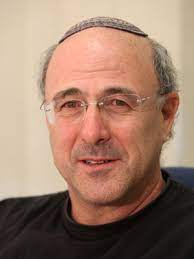}}]{Michael Werman}
Michael Werman received his Ph.D. degree from The Hebrew University, in 1986, where he is currently a Professor of Computer Science. His current research interests include designing computer algorithms and mathematical tools for analyzing, understanding, and synthesizing pictures.
\end{IEEEbiography}

\section{Appendix}\label{appendix}

\end{document}